\documentclass[journal]{IEEEtran}

\ifCLASSINFOpdf

\else

\fi

\usepackage{epsfig} % added for .eps graphics extension
\usepackage{graphicx}

\graphicspath{{graphics/}}

\usepackage{epstopdf}

\usepackage{subcaption} %had to add this so my graphics would show up
\usepackage{multirow}
\usepackage{lscape}

\usepackage{lineno,hyperref}
 \usepackage{flushend}

\usepackage[table]{xcolor}% http://ctan.org/pkg/xcolor
\usepackage{color, colortbl}
\definecolor{BlueTable}{rgb}{0.30,0.58,0.93}
\definecolor{lightGray}{rgb}{0.8,0.8,0.8}

\usepackage{rotating} % For the sideways table
\modulolinenumbers[30]

\begin{document}

\title{Analyzing Covariate Influence on Gender and Race Prediction from Near-Infrared Ocular Images}
%\title{Predicting Gender and Race from Near Infrared Iris and Periocular Images}

\author{
\IEEEauthorblockN{Denton Bobeldyk and Arun Ross}
{\\\small{Preprint title: Predicting Gender and Race from Near Infrared Iris and Periocular Images}}
}

\maketitle
\thispagestyle{empty}

\begin{abstract}

Recent research has explored the possibility of automatically deducing information such as gender, age and race of an individual from their biometric data. While the face modality has been extensively studied in this regard, the iris modality less so. In this paper, we first review the medical literature to establish a biological basis for extracting gender and race cues from the iris. Then, we demonstrate that it is possible to use simple texture descriptors, like BSIF (Binarized Statistical Image Feature) and LBP (Local Binary Patterns), to extract gender and race attributes from an NIR ocular image used in a typical iris recognition system. The proposed method predicts gender and race from a \emph{single} eye image with an accuracy of 86\% and 90\%, respectively. In addition, the following analysis are conducted: (a) the role of different parts of the ocular region on attribute prediction; (b) the influence of gender on race prediction, and vice-versa; (c) the impact of eye color on gender and race prediction; (d) the impact of image blur on gender and race prediction; (e) the generalizability of the method across different datasets; and (f) the consistency of prediction performance across the left and right eyes.
\end{abstract}

\IEEEpeerreviewmaketitle

\begin{figure*}
\centering
\begin{tabular}{c c c}
\captionsetup{type=figure}
\includegraphics[width=0.20\textwidth]{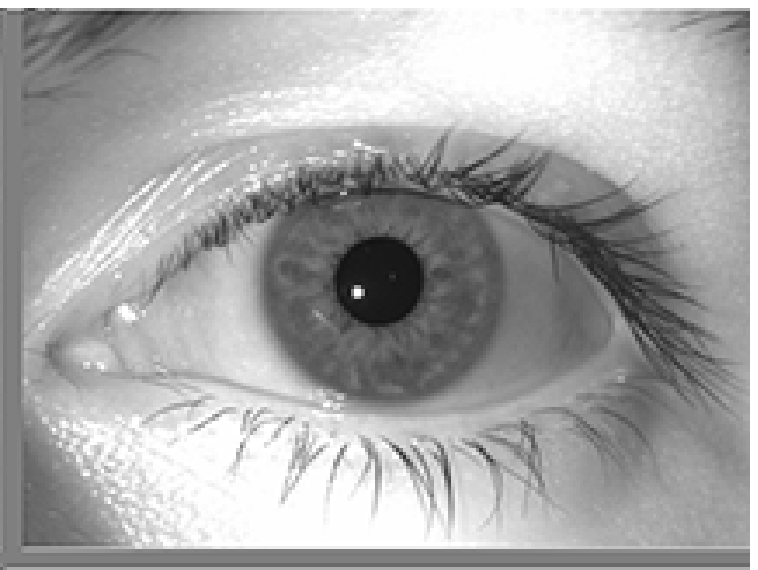}&\includegraphics[width=0.15\textwidth]{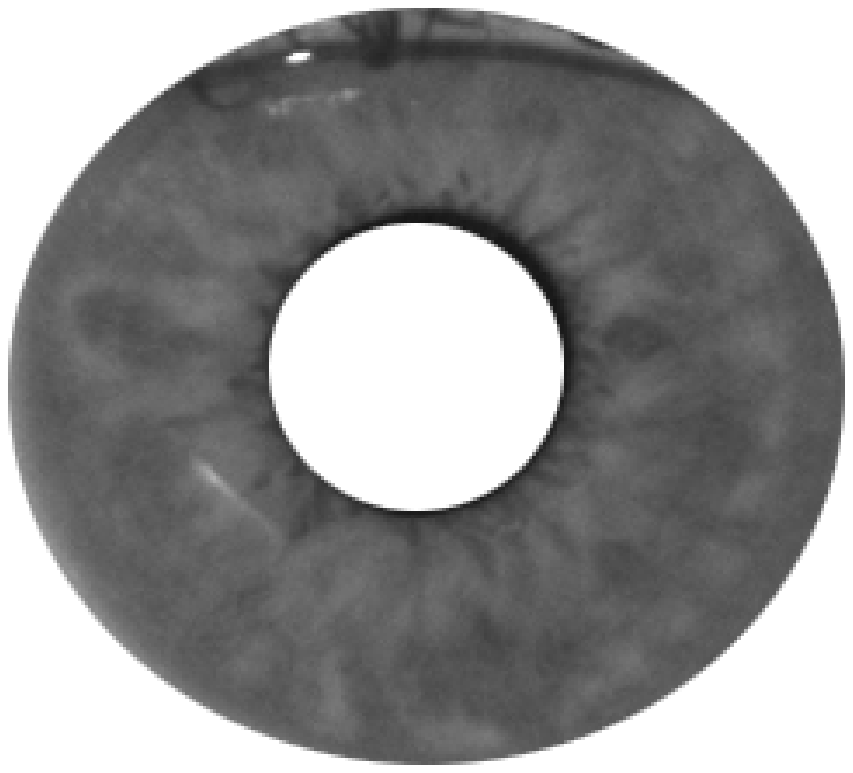}&\includegraphics[width=0.15\textwidth]{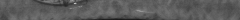} \\
(a) Ocular image & (b) Segmented iris region & (c) Normalized Iris\\
\end{tabular}
\captionof{figure}{The process of iris recognition typically involves (a) imaging the ocular region of the eye using an NIR camera, (b) segmenting the annular iris region from the ocular image, and (c) unwrapping the annular iris region into a fixed-size rectangular entity referred to as normalized iris. Image (a) is from~\cite{doyle2013variation}.}

\label{irisSegmentation}
\end{figure*}

\section{Introduction}

\subsection{Iris Recognition System}
A biometric system utilizes the physical or behavioral traits of an individual to automatically recognize an individual~\cite{jain2011introduction}. Examples of such traits include face, fingerprint, iris, voice and gait. The focus of this paper is on the iris biometric trait. The iris is the colored and richly textured annular region of the eye surrounding the pupil (see Figure~\ref{irisSegmentation}). The rich texture of dark colored irides is not easily discernible in the visible wavelength; therefore, the iris is typically imaged in the Near Infrared (NIR) spectrum since longer wavelengths tend to penetrate deeper into the multi-layered iris structure thereby eliciting the texture of even dark-colored eyes. Further, the NIR image acquisition process does not excite the pupil, thereby ensuring that the iris texture is not unduly deformed due to pupil dynamics~\cite{clark2013theoretical}.

\subsection{Predictable Attributes from NIR Ocular Images}

In addition to performing the task of recognizing an individual~\cite{daugman2004iris}, it is possible to predict attributes about the individual, such as gender, race and age, from the raw biometric data itself. These attributes are referred to as soft biometrics~\cite{dantcheva2015else}. Soft biometric attributes may not be discriminative enough to uniquely identify an individual, but can be used to increase the recognition accuracy of a biometric system~\cite{jain2004can}. In addition to increased performance, there are several other motivating factors to glean these attributes from the raw biometric data. Firstly, databases containing biometric data could be automatically processed and the soft biometric information aggregated (e.g., proportion of Asian Males in a database). The aggregated information could then be used for statistical descriptions of the database or to index the database for faster retrieval of identities. Secondly, a semantic description of an individual (e.g., "Middle-aged Caucasian Male") can be automatically generated from a single biometric sample that could prove useful in forensic applications to either reduce the list of potential matching candidates or to exclude a suspect from the search. Thirdly, soft biometric attributes may be beneficial in scenarios where the input image does not lend itself to the identity recognition task. For example, out-of-focus iris images may result in poor recognition accuracy while still allowing for the extraction of attributes such as gender and race. Fourthly, soft biometric attributes can potentially enable cross-spectral recognition, where images acquired in the NIR spectrum have to be compared against their visible spectrum counterparts~\cite{jillela2014matching}.

The work discussed in this paper will focus specifically on the prediction of gender\footnote{The terms `gender' and `sex' have been used interchangeably in the biometric literature. There is, however, a specific definition provided by the Health and Medicine Division of the National Academies of Science, Engineering and Medicine. They state that sex is biologically or genetically determined, while gender is culturally determined~\cite{torgrimson2005sex}.} and race\footnote{The terms `ethnicity' and `race' have been used interchangeably in related biometric literature. An exact definition of either of these two terms appears to be debatable, and further information can be found in \cite{billinger2007another}.}  from NIR ocular images typically used in iris recognition systems. Some sample NIR ocular images with the race and gender label are displayed in Figure~\ref{sampleAttributedOcularImages}. The prediction of attributes from other biometric traits has also been actively studied in the literature (see Table \ref{nonIrisAttributePrediction}). There is also related attribute prediction research from visible spectrum ocular images~\cite{lyle2010soft}, \cite{rattani2017gender}, \cite{tapia2017genderMultispectral} and \cite{merkow2010exploration}.

\begin{table*}
\caption{Examples of attribute prediction using different biometric traits.}
\label{nonIrisAttributePrediction}
\centering
\resizebox{\textwidth}{!}{\begin{tabular}{l l l l l l}
\hline\hline
\multirow{2}{*}{\textbf{Trait}}& \multirow{2}{*}{\textbf{Attribute}} &\multirow{2}{*}{\textbf{Method Used}} &\textbf{Dataset}& \textbf{Prediction} &\multirow{2}{*}{Reference}\\
&& &\textbf{(\#images/\#subjects)}& \textbf{Accuracy} &\\

\hline
Body&Gender&Figure Sequential with SVM&HumanID (100 Subjects)&96.7\% &\cite{yoo2005gender} \\
Face&Age&LBP, HOG, Bio Inspired Features&YGA (8000 images)&94.9\% &\cite{guo2012human}\\
&&with a nonlinear SVM&& &\\

NIR Face& Gender & LBP with SVM&CBSR NIR (3200 images) &93.59\%&\cite{ross2011can}\\
Fingerprint & Gender &Discrete Wavelet Transform, & Private (498 images)&96.59\%&\cite{nagabhyruSnehaDissertation} \\
 &  &Wavelet Analysis &  && \\

Face&Ethnicity&2D and 3D Multi Scale Multi &FRGCv2.0 (180 subjects)&99.5\%&\cite{zhang2009multimodal}\\
&& Ratio LBP with Adaboost &&&\\

\hline
\hline
\end{tabular}}
\end{table*}

\subsection{Ocular Anatomy}
\label{ocularAnatomy}

\begin{figure*}[t]   % ocular anatomy
\center
\includegraphics[width = 3in]{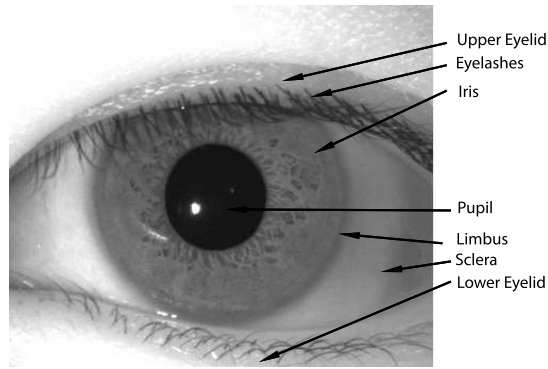}
    \caption{An NIR image depicting the various parts of the ocular region. }
    \label{labeledOcularImage}
\end{figure*}

A discussion of the ocular anatomy is useful in understanding the type of gender and race markers present in the ocular region. The ocular region could be defined as the region housing the eye (see Figure \ref{labeledOcularImage}). The eyeball has both upper and lower eyelids that provide a protective and lubricative function to the eyeball. The upper eyelid contains the levator palpebrae superioris, which is the muscle that allows the eye to blink \cite{ansari2016atlas}. The gap between the upper and lower eyelid is the palpebral fissure. The iris and pupil region are located between the upper and lower eyelids.

Previous research has established the distinctiveness of the iris patterns of an individual \cite{daugman2003importance}. The iris texture is imparted by an agglomeration of several anatomical features: stroma, Fuchs' crypts, Wolfflin nodules, pigmentation dots, and contraction furrows, to name a few. The anterior portion of the iris has two distinct zones, the pupillary zone and the ciliary zone, that are separated by the collarette. There are some correlations between features that are present in the iris. For example, an iris that has no Fuchs' crypts may have clearly distinguishable contraction furrows \cite{sturm2009genetics}. A decrease in the density of the stroma has been observed as the number of Fuchs' crypts increases. As the density decreases, the contraction furrows has been observed to decrease.

\begin{table*}[t]
\center
\begin{tabular}{c c c c}
\includegraphics[width=0.15\textwidth]{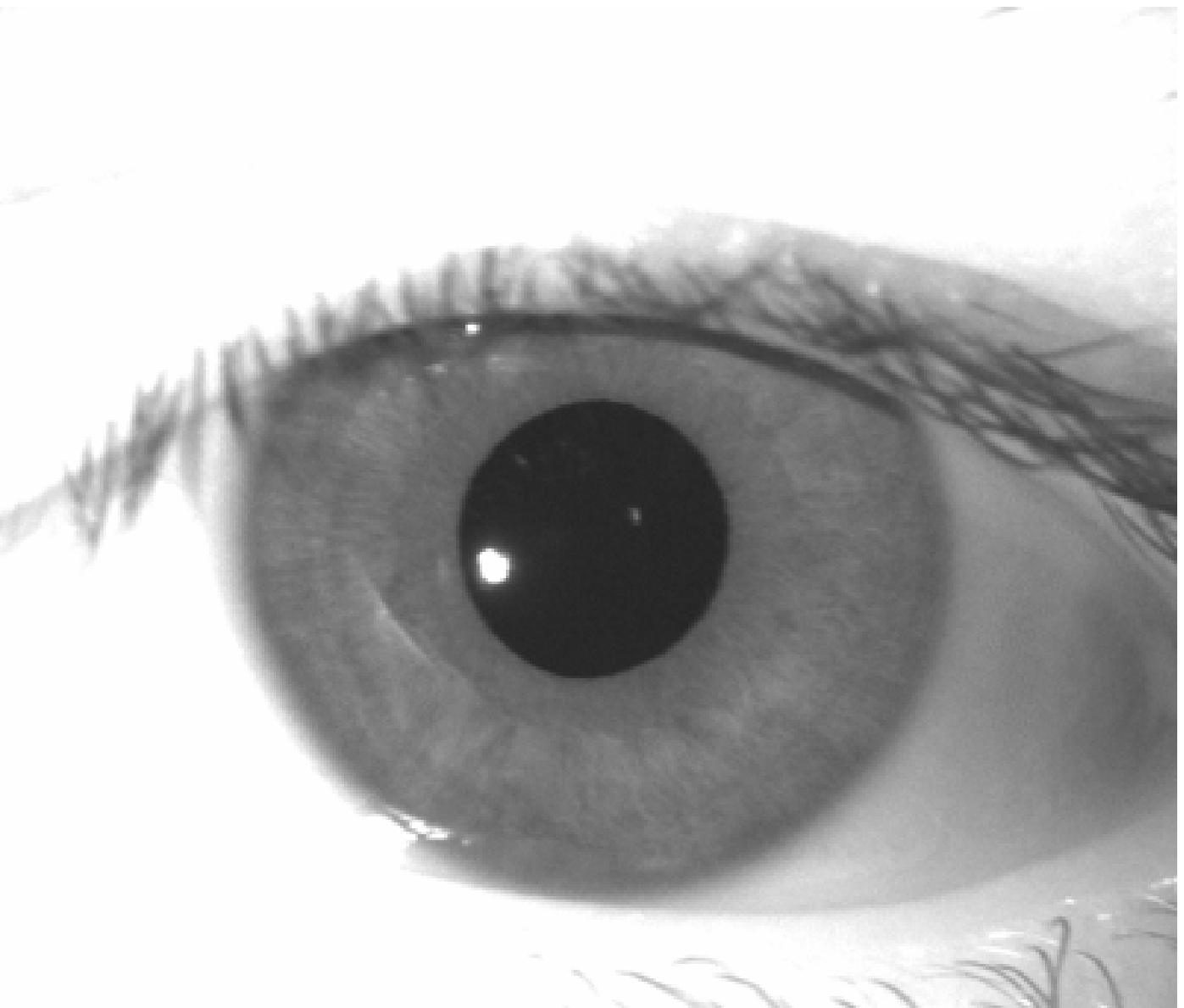}&
\includegraphics[width=0.15\textwidth]{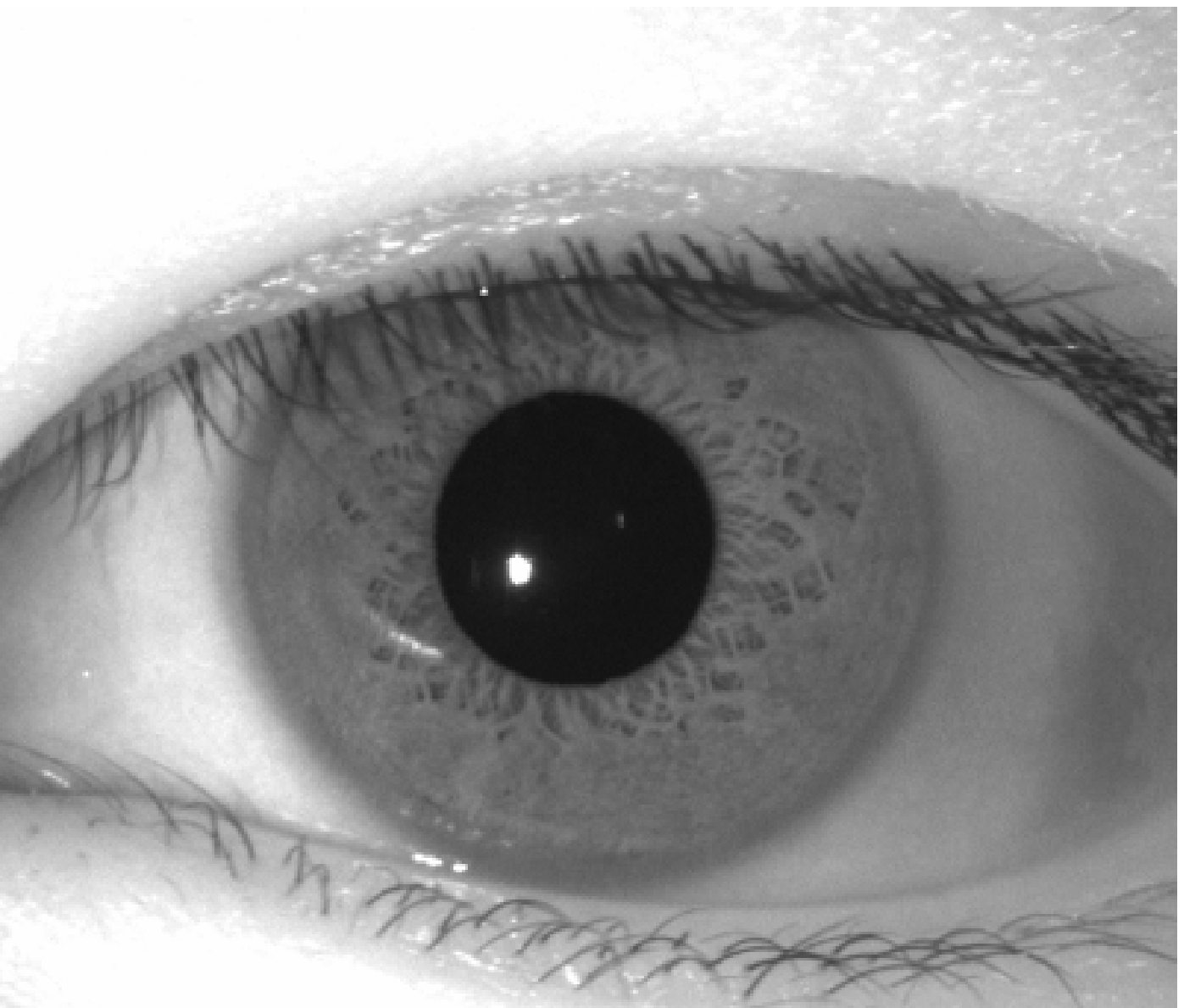}&
\includegraphics[width=0.15\textwidth]{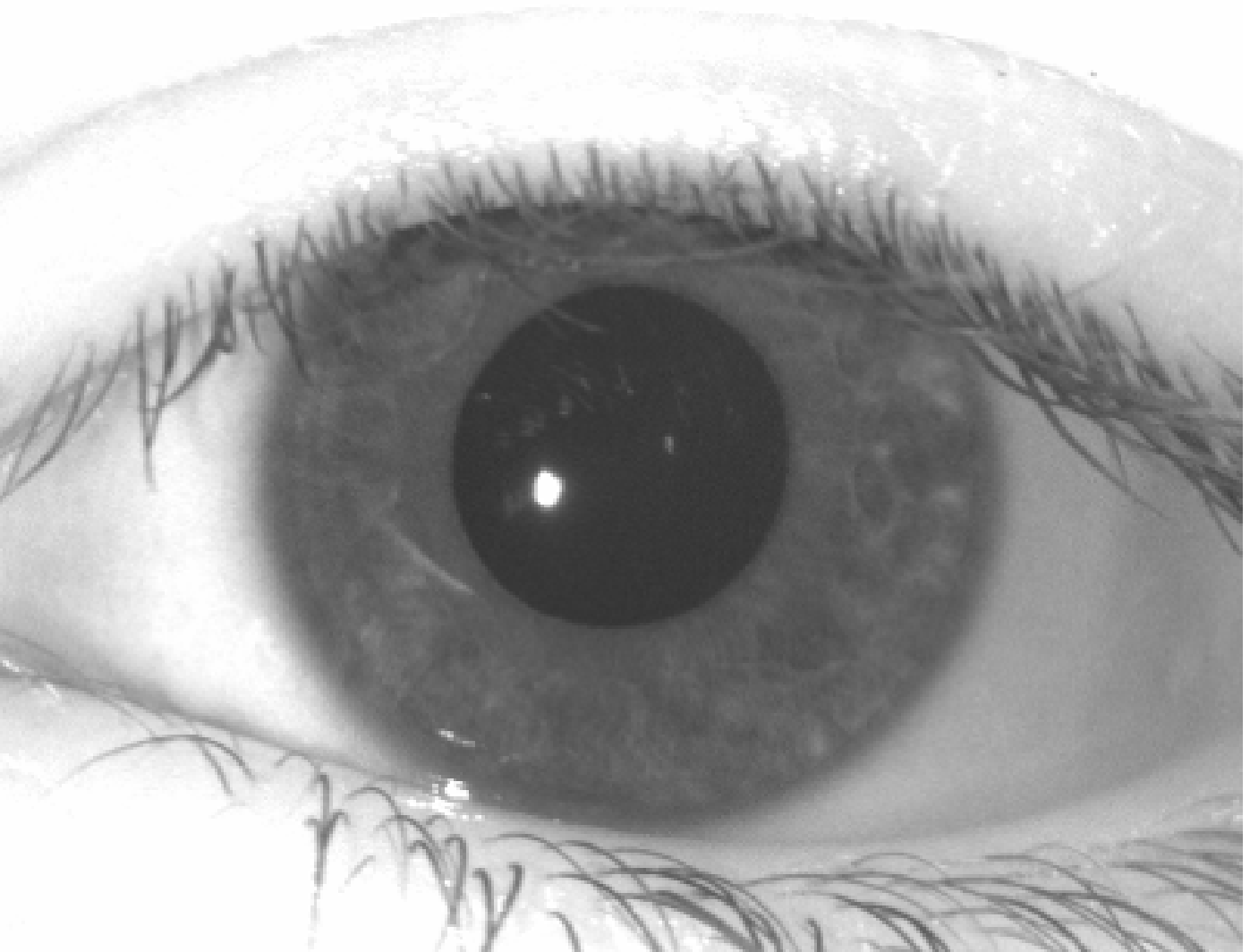}&
\includegraphics[width=0.15\textwidth]{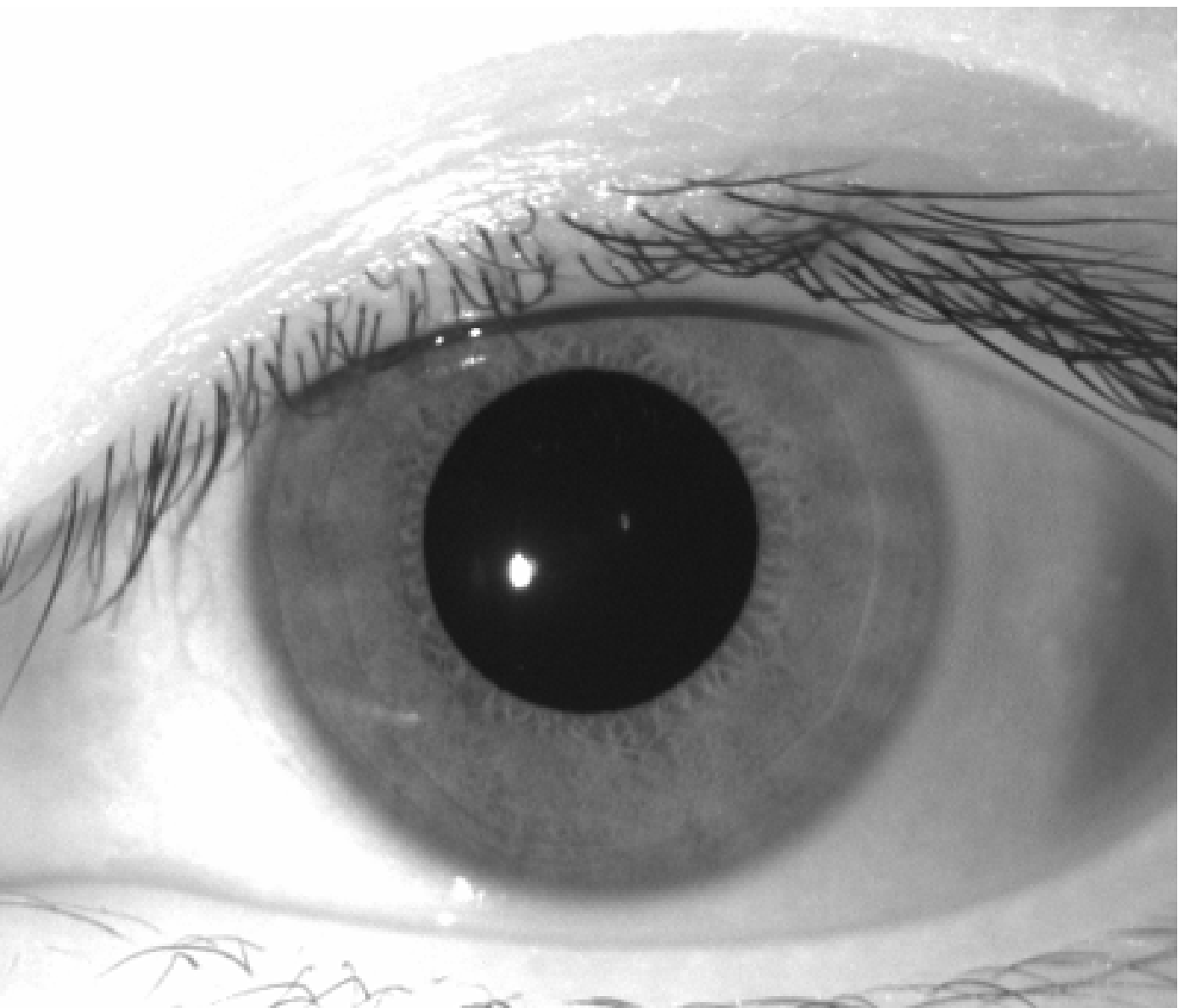} \\
\end{tabular}
\captionof{figure}{Examples of ocular images pertaining to different categories of individuals. From Left to Right: male Caucasian, male Non-Caucasian, female Caucasian, female Non-Caucasian. The images are from \cite{doyle2013variation}.}
\label{sampleAttributedOcularImages}
\end{table*}

The medical literature suggests both geometric and textural difference between male and female irides. From a textural perspective, Larsson and Pederson \cite{larsson2004genetic} found that males have a greater number of Fuchs' crypts than females. From a geometric perspective, Sanchis et. al \cite{sanchis2012white} report that the pupil diameters are greater in emmetropic females.\footnote{The emmetropic state of the subjects in our dataset is unknown. An experiment was conducted to determine if there was a statistically significant difference in pupil diameter between males and females in the dataset that was used. Using the diameter of the iris as determined by a COTS software,  there was no statistically significant gender-specific difference found in either the left (male \{$\mu = 80.66, \sigma = 16.2$\}, female \{$\mu = 80.9, \sigma =16.3$\}) or right (male \{$\mu = 79.6, \sigma = 15.5$\}, female \{$\mu = 79.7, \sigma =16.2$\}) eyes.}

If the entire ocular region is considered (and not just the iris), it has been found that the lacrimal glands of men are 30\% larger and contain 45\% more cells than those of females \cite{wagner2008sex}. There are also significant corneal differences in that women have steeper corneas\footnote{If we liken a cornea to a sine wave, we can think of a steep cornea as a sine wave with a higher amplitude}  than men and their corneas are also thinner \cite{suzuki2006influence}\cite{wagner2008sex}. Other differences in the cornea include ``diameter, curvature, thickness, sensitivity and wetting time of the cornea'' \cite{suzuki2006influence}. While a number of the aforementioned features may not manifest themselves in a 2D NIR ocular image,  we hypothesize that the texture of the ocular region, including the iris, may offer gender (or sex) cues of an individual.

There exists textural differences in the iris between races as well. Edwards et al. \cite{edwards2016analysis} examined images of irides in the visible spectrum from 3 separate populations: South Asian, East Asian and European. Europeans were found to have a higher grade\footnote{In their work, the authors defined 4 categories of Fuchs' crypts. Category 1 contains no crypts, while category 4 contains `at least three large crypts located in three or more quadrants of the iris'  \cite{edwards2016analysis}. } of Fuchs' crypts, more pigment spots, more extended contraction furrows, and more extended Wolfflin nodules than East Asians \cite{edwards2016analysis}. East Asians were found to have a lower grade of Fuchs' crypts than both Europeans and South Asians. Europeans had the largest iris width, followed by South Asians, and then by East Asians \cite{edwards2016analysis}. As for eye color, East Asians had the darkest while Europeans had the lightest.

\subsection{Paper Contributions}

This article focuses on the covariate influence of predicting race and gender from the ocular region using simple texture descriptors.  Thus, the contributions of the paper are as follows:

\begin{itemize}

\item The prediction accuracies due to the iris-only region is contrasted with that of the extended ocular region for both race and gender using multiple texture descriptors.
\item Determine whether eye location (i.e., left or right eye) has a significant impact on gender and race prediction.
\item Determine whether Caucasian or Non-Caucasian subjects exhibit higher gender prediction accuracy
\item Determine whether Male or Female subjects exhibit higher race prediction accuracy
\item Study the sensitivity of gender and race prediction to image blur.
\item Study the impact of eye color on gender and race prediction.

\end{itemize}

The rest of the paper is organized as follows: Section \ref{relatedWorkSection} discusses related work in the field. Section \ref{featureExtractionSection} discusses the proposed feature extraction method. Section \ref{datasetsSection} describes the datasets that were used. Section \ref{experimentsSection} discusses the experiments conducted and the associated results. Section \ref{discussionSection} discusses the experimental results and summarizes the findings of this work. Section \ref{futureWorkSection} provides a list of future directions for this research.

\section{Related Work}
\label{relatedWorkSection}

\label{preambleRelatedWork}
The problem of attribute prediction, be it gender or race, is typically posed as a  pattern classification problem where a feature set extracted from the biometric data (e.g., an ocular image) is input to a classifier (e.g., SVM, decision tree, etc.)  in order to produce the attribute label (e.g., `Male'). The classifier itself is trained in a supervised manner with a training set consisting  of ocular data labeled with attributes. The performance of the prediction algorithm is then evaluated on an independent test set. Good practice~\cite{jain2015guidelines} dictates that the subjects in the training set and test set be mutually exclusive. An optimally biased predictor can be produced if there is an overlap of subjects in the training and test sets as indicated in \cite{bobeldyk2016iris,tapia2016gender}. While most recent work in attribute prediction from iris have clearly adopted a subject-disjoint protocol~\cite{bobeldyk2016iris,tapia2016gender, tapia2018gender, bobeldyk2018predicting}, some of the earlier papers on this topic have been ambiguous on this front~\cite{thomas2007learning,bansal2012, qiu2006global, qiu2007learning}. Table \ref{relatedWorkGenderAccuraciesSingle} and Table \ref{relatedWorkRaceAccuracies}, respectfully, summarize the previous work on gender and race prediction from a single NIR image. Table \ref{relatedWorkGenderAccuraciesFused} summarizes the gender prediction work that use images from both the left and right eyes in a fusion framework.

\subsection{Gender}
\label{genderRelatedWork}

%\begin{landscape}
\begin{table*} % Gender prediction related work table
\caption{Gender Prediction - Related Work (Left or Right Eye Image).\\{\tiny Work that uses the publicly available GFI dataset are highlighted for ease of comparison}}
%\caption{Gender Prediction - Related Work (Single Image).\\{\tiny Work that uses the publicly available GFI dataset~\cite{tapia2016gender} are highlighted for ease of comparison}}
\label{relatedWorkGenderAccuraciesSingle} % This must come after the caption
\centering
\resizebox{0.99\textwidth}{!}{\begin{tabular}{c c c c c c c c c}
\hline\hline
 \multirow{2}{*}{\textbf{Authors}} &\textbf{Year}& \textbf{Subject-Disjoint} &\textbf{Dataset}&\textbf{Number}&\textbf{Number}&\textbf{Features}&\textbf{Prediction} \\
 & & \textbf{Specified} & &\textbf{of Subjects} &\textbf{of Images} & &\textbf{Accuracy}\\
 \hline
Thomas et al. \cite{thomas2007learning}  &2007& No & Private & Unknown & 57,137 & Geometric/Texture features&80\% \\
Bansal et al. \cite{bansal2012}  & 2012& No  & Private & 150 & 300 & Statistical/Texture features&83.06\%\\
\hline
%\multirow{2}{*}{Singh et al. \cite{singh2017gender}}&\multirow{2}{*}{2017} &No&ND-Iris-0405&356&60,259&Deep class-&82.53\%\\
Singh et al. \cite{singh2017gender}&2017&No&ND-Iris-0405&356&60,259&Deep class-encoder&82.53\%\\
 \rowcolor{lightGray}
Singh et al. \cite{singh2017gender}&2017&Yes&GFI&1500&3000&Deep class-encoder&83.17\%\\
\hline
Lagree \& Bowyer \cite{lagree2011predicting}  & 2011& Yes  & Private &120&1200&Basic texture filters&62\%\\
Fairhurst et al.~\cite{da2015exploring}&2015&Yes &BioSecure&200&1600&Geometric and Texture Features&81.43\%\\
Bobeldyk \& Ross \cite{bobeldyk2016iris} &2016& Yes  & Private & 1083 & 3314 & BSIF & 85.7 \%\\
\rowcolor{lightGray}
Kuehlkamp et al. \cite{kuehlkamp2017gender}&2017& Yes & GFI & 1500 & 3000 & CNN and MLPs & ~80\%\\
Tapia et al.~\cite{tapia2017genderMultispectral}&2017&Yes&CROSS-EYED&120& 1920 &HOG w/ feature selection &90.0\%\\
\rowcolor{lightGray}
This Work &2018& Yes  & GFI &1500& 3000 & BSIF & \textcolor{blue}{84.4\%}\\
This Work &2018& Yes  & BioCOP2009 & 1096 & 41,780 & BSIF & 86.0\%\\
\hline\hline
\end{tabular}}
%\end{tabular}
\end{table*}

\begin{table*} % Gender prediction related work table
\caption{Gender Prediction - Related Work (Left + Right Eye Fusion)}
\label{relatedWorkGenderAccuraciesFused} % This must come after the caption
\centering
\resizebox{0.99\textwidth}{!}{\begin{tabular}{c c c c c c c c c}
\hline\hline
 \multirow{2}{*}{\textbf{Authors}} &\textbf{Year} &\textbf{Subject-Disjoint} &\textbf{Dataset}&\textbf{Number}&\textbf{Number}&\textbf{Features}&\textbf{Prediction} \\
 & & \textbf{Specified} & &\textbf{of Subjects} &\textbf{of Images} &&\textbf{Accuracy}\\
 \hline
Tapia et al. \cite{tapia2014gender}& 2014& No  & GFI & 1500\protect\footnotemark & 3000 & Uniform LBP &91.33\%\\
Tapia et al. \cite{tapia2016gender} & 2016&Yes  & GFI & 1500 & 3000 &  Iriscode and weighted feature selection &89\% \\
Tapia et al. \cite{tapia2017gender} & 2017& Yes & GFI & 1500 & 3000 & CNN fusing of separate left/right CNNs & 84.66\%\\
Tapia \& Aravena \cite{tapia2018gender} & 2018 & Yes & GFI& 1500 & 3000 & CNN (Reduced version of LeNet)&  87.26\% \\
\hline\hline
\end{tabular}}
%\end{tabular}
\end{table*}

\footnotetext{The published paper claims 1500 subjects; however it was discovered during our experiments that there was actually far less number of subjects. The authors confirmed their error via email and in one of their subsequent publications \cite{tapia2016gender}.}

One of the earliest work in prediction of gender from the iris was published by Thomas et al. \cite{thomas2007learning}. The authors assembled a dataset of $57,137$ ocular images. The iris was extracted from each of the ocular images and normalized into a $20\times240$ image using Daugman's rubber sheet method~\cite{daugman2003importance}. A feature vector was generated from the normalized image by applying a one dimensional Gabor filter. Feature selection was performed using an information gain metric. The resulting reduced feature vector was classified as `Male' or `Female' by a decision tree algorithm. Only left iris images were used for their experiments. The authors were able to achieve an accuracy that was `upwards of  80\% with Bagging' \footnote{It was not stated whether a subject-disjoint training and test set were used} using only the Caucasian subjects in their dataset.\footnote{A prediction accuracy of 75\% was achieved using all of the images}

Bansal et al. \cite{bansal2012}  were able to achieve an 83.06\% gender  classification accuracy using statistical and wavelet features along with an SVM classifier. Occlusions from the iris region were removed (i.e., eyelids, eyelashes) using a masking algorithm. The size of their dataset, however, was quite small with only 150 subjects and 300 iris images. 100 of the subjects were male and 50 of the subjects were female. However, it is not clear if they used a subject-disjoint evaluation protocol.

Lagree and Bowyer performed gender classification on a dataset of 600 iris images each of which was normalized to a $40\times240$ rectangular image. 8 horizontal regions of 5-pixel width and 10 vertical regions of 24-pixel width were then created. Using the created regions and some simple texture filters (i.e., for detecting spots, lines), an 882-dimensional feature vector was computed. An SVM classifier was then applied (specifically the WEKA SMO algorithm) for classification, achieving a 62\% accuracy.

Tapia et al. \cite{tapia2016gender} extended their earlier work \cite{tapia2014gender} to determine if the iriscode\footnote{The iriscode is a feature vector commonly used in iris recognition. Additional details can be found in \cite{daugman1994biometric}.} contained sex predictive information. Using an SVM classifier on the iriscode they obtained a $77.3\%$ accuracy for the left iris and a $74.66\%$ accuracy for the right iris. Their previous work in \cite{tapia2014gender} did not use a subject-disjoint training/testing dataset resulting in an `accuracy of $87.33\%$ for the left iris and $84.66\%$ for the right iris.' The accuracy was artificially high due to a labeling error in the dataset which did not allow for a subject-disjoint training/testing dataset~\cite{tapia2016gender}.  The method in~\cite{tapia2016gender} improved upon their results by applying fusion between the left and right iriscodes using Weighted Conditional Mutual Information Maximization fusion which resulted in an $89.00 \pm 0.68\%$ prediction accuracy. Tapia et al. \cite{tapia2017gender} continued their work on deducing gender from iris by utilizing a CNN architecture that fused \emph{normalized} iris images from the left and right eyes and were able to achieve an 84.66\% accuracy. Tapia et al. \cite{tapia2017gender} cited the small size of their dataset  as a possible reason for their performance not surpassing that of their previous work \cite{tapia2016gender}. In \cite{tapia2018gender}, Tapia \& Aravena proposed a CNN architecture that fused the left and right \emph{periocular} NIR images together. The model utilized three CNNs: one for the left eye, one for the right eye and one to fuse the left and right eye models together. They were able to achieve an 87.26\% prediction accuracy.

\begin{figure*}
\begin{center}  %Ocular region vs iris-only, vs. iris-excluded
\begin{tabular}{c c c}
\centering
\label{differentRegions}
\centering
\captionsetup{type=figure}  % Including this to remove the latex warning for using 'captionof'
\includegraphics[width=0.20\textwidth]{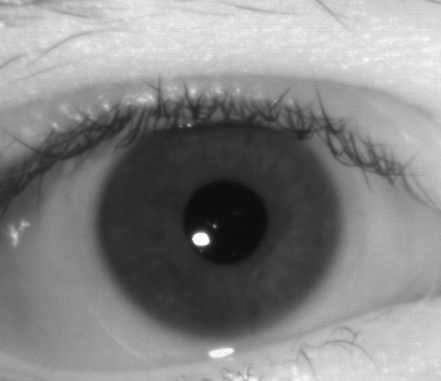}&\includegraphics[width=0.12\textwidth]{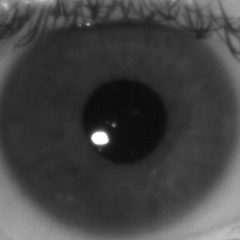}&\includegraphics[width=0.20\textwidth]{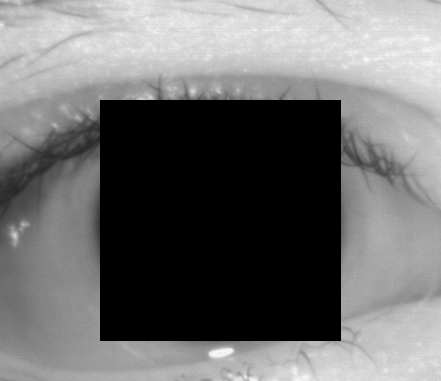} \\
\fontsize{7}{9}\selectfont(a) (Extended) Ocular Image &\fontsize{7}{9}\selectfont(b) Iris-Only Image &\fontsize{7}{9}\selectfont (c) Iris-Excluded Ocular Image \\
\end{tabular}
\captionof{figure}{The three different regions in an NIR ocular image that are independently considered for the gender and race prediction tasks. Images are from \cite{doyle2013variation}.}
\label{irisRegionExamples}
\end{center}
\end{figure*}

Most biometric recognition work pertaining to NIR iris images have focused on extracting the iris region from the captured ocular image (see Figure \ref{irisSegmentation}). Thus, algorithms for soft biometric prediction have typically focused on the iris region rather than the extended ocular region (see Figure \ref{irisRegionExamples}). Recent work \cite{bobeldyk2016iris} based on the Binarized Statistical Image Feature (BSIF) descriptor has shown that the extended ocular region commonly imaged by iris recognition systems provides greater sex prediction accuracy than the iris-only region. Predicting soft biometric attributes from the ocular region provides one major advantage over the iris region in that it does not require a potentially error prone algorithm for iris region extraction. Bobeldyk and Ross \cite{bobeldyk2016iris} were able to achieve an 85.7\% sex prediction accuracy using concatenated histograms from tesselated regions of the BSIF code computed from NIR ocular images.

Fairhurst et al.~\cite{da2015exploring} utilized geometric features from the ocular image and texture features from the normalized iris image and were able to achieve an 81.43\% prediction accuracy on a subset of the BioSecure dataset consisting of 200 subjects and 1600 images.

Singh et al. \cite{singh2017gender} use a variant of an auto-encoder that includes the attribute class label alongside the reconstruction layer. They used NIR ocular images that were resized to $48\times64$ pixels. Their proposed method was tested on both the GFI and ND-Iris-0405 datasets from Notre Dame. The experiments on the GFI dataset utilized the 80-20 subject-disjoint split specified in the dataset. While experiments using the ND-Iris-0405 dataset were not indicated as being subject-disjoint, their paper states: `All protocols ensure mutually exclusive training and testing sets, such that there is no \textit{image} [emphasis added] which occurs in both the partitions'.

Kuehlkamp et al. \cite{kuehlkamp2017gender} studied the effect of mascara on predicting gender from iris. Using only the occlusion mask from each of the images, they achieved a 60\% gender prediction accuracy. They went on to show that LBP combined with an MLP network was able to achieve a 66\% accuracy. Using the entire `eye' image they were able to achieve around 80\% using CNNs and MLPs.

Tapia et al.~\cite{tapia2018gender} used a feature selection model that was similar to their earlier work on iris~\cite{tapia2016gender}, but applied to periocular images. They were able to achieve a 90.0\% prediction accuracy on a dataset containing 120 subjects and 1920 images.

Previous work utilizing a \emph{single} eye image (left or right) are displayed in Table~\ref{relatedWorkGenderAccuraciesSingle} and those utilizing a \emph{fused} model combining the left and right are shown in Table~\ref{relatedWorkGenderAccuraciesFused}. For ease of comparison, the works that utilized the publicly available GFI dataset are highlighted in Table~\ref{relatedWorkGenderAccuraciesSingle}. The reported accuracy for our current work on the GFI dataset is based on a prediction model that was trained on the BioCOP2009 dataset (see Section~\ref{genderCrossDatasetSection}).

It should also be noted that there are some gender prediction work using the periocular region in the \emph{visible wavelength spectrum} in \cite{lyle2010soft}, \cite{rattani2017gender}, \cite{tapia2017genderMultispectral} and \cite{merkow2010exploration}.

\begin{table*} %\label{relatedWorkEthnicityAccuracies}
\caption{Race Prediction - Related Work}
\label{relatedWorkRaceAccuracies} % This must come after the caption
\centering
\resizebox{\textwidth}{!}{\begin{tabular}{c c c c c c c}

\hline\hline
 \multirow{2}{*}{\textbf{Authors}} & \textbf{Subject Disjoint} & \textbf{Dataset Used} & \textbf{\# of subjects} & \textbf{\# of images}& \textbf{Features Used}& \textbf{Prediction} \\
 & \textbf{Specified} & & & & & \textbf{Accuracy}\\
 \hline
Qiu et al. \cite{qiu2006global} & No&CASIA, UPOL, UBIRIS&Unknown &3982&Gabor filters & $85.95\%$ \\
Qiu et al. \cite{qiu2007learning} & No&Proprietary &60&2400&Gabor filters & $91.02\%$ \\
Singh et al. \cite{singh2017gender} & No&ND-Iris-0405/Multi-Ethnicity&240/Unknown   &60,259/60,310&Deep class-encoder&94.33\%/97.38\%\\
\hline
Lagree \& Bowyer \cite{lagree2011predicting}  & Yes &Proprietary &120 &1200&Basic texture filters & $90.58\%$ \\
Proposed Work & Yes &BioCOP2009 &1096& 41780 &BSIF & $90.1\%$\\
\hline\hline
\end{tabular}}
\end{table*}

\subsection{Race}
There are only a few papers that attempt to deduce race from NIR iris images.  In \cite{qiu2006global} and \cite{qiu2007learning} the authors do not state whether their train and test partitions are subject-disjoint, and the size of the datasets are quite small (3982 and 2400 images, respectively). In both publications, Qiu et al. \cite{qiu2006global, qiu2007learning} utilized the texture generated from Gabor filters to create a feature vector that was classified using AdaBoost and SVM (respectively) classifiers. A smaller region within the captured iris image was used in order to minimize occlusions from eyelids or eyelashes.

Singh et al. \cite{singh2017gender} also did not specify a subject-disjoint experimental protocol. Their proposed method used a variant of an auto-encoder that includes the class label alongside the reconstruction layer. The experiments were performed on the ND-Iris-0405 dataset as well as a multi-ethnicity iris dataset composed of three separate datasets. Each class (Asian, Indian, Caucasian) was represented by a distinct dataset. They achieved a 94.33\% prediction accuracy on the ND-Iris-0405 dataset and 97.38\% on the multi-ethnicity iris dataset. However, it is not clear if the multi-ethnicity results were optimistically biased due to the use of different datasets for the 3 classes. As pointed out by El Naggar and Ross \cite{Ross_WIFS_15}, dataset-specific cues are often present in the images.

To the best of our knowledge, Lagree and Bowyer \cite{lagree2011predicting} were the first to look at the impact of gender on race prediction. Race was predicted by training and testing on images from only female subjects, as well as images from only male subjects. They were able to achieve an 82.8\% race prediction accuracy using only female subjects, and a 92\% accuracy using only male subjects. Using all of the images (both male and female) their best prediction accuracy was 87.6\%. Using a slightly larger dataset, they achieved a 90.58\% accuracy for a mixed male and female dataset.

\section{Feature Extraction}
\label{featureExtractionSection}

One of the goals of our work is to establish the utility of simple texture descriptors for attribute prediction. Uniform local binary patterns (LBP) \cite{ojala2002multiresolution} and binarized statistical image features (BSIF) are two texture descriptors that have performed well on the Outex and Curet texture datasets~\cite{kannala2012bsif}. Both have shown to perform well in the attribute prediction domain~\cite{bobeldyk2016iris, tapia2014gender}, with BSIF outperforming LBP in both domains (texture and attribute prediction). Three texture descriptors were considered in this work: BSIF, LBP and LPQ (Local Phase Quantization).

LBP~\cite{ojala2002multiresolution} encodes local texture information by comparing the value of every pixel of an image with each of its respective neighboring pixels. This results in a binary code whose length is equal to the number of neighboring pixels considered. The binary sequence is then converted into a decimal value, thereby generating an LBP code for every pixel in the image.

LPQ~\cite{ojansivu2008blur} encodes local texture information by utilizing the phase information of an image. A sliding rectangular window is used, so that at each pixel location, an 8-bit binary code is generated utilizing the phase information from the 2-D Discrete Fourier Transform. A histogram of those generated values results in a 256-dimensional feature vector.

BSIF was introduced by Kanala and Rahtu \cite{kannala2012bsif} as a texture descriptor. BSIF projects the image into a subspace by convolving the image with pregenerated filters. The pregenerated filters are created from 13 natural images supplied by the authors of~\cite{hyvarinen2009natural}. 50,000 patches of size $k\times k$ are randomly sampled from the 13 natural images. Principal component analysis is applied, keeping only the top \textbf{n} components. Independent component analysis is then applied generating \textbf{n} filters of size $k \times k$. The authors of  \cite{kannala2012bsif} provide the pregenerated filters for $k=\{3,5,7,9,11,13,15,17\} $ and $n = \{5-12\}$.\footnote{For $n=\{9-12\}$, $k=3$ was not made available by~\cite{kannala2012bsif}.}

Each of the \textbf{n} pregenerated filters are convolved with the image and the response is binarized. If the response is greater than zero, a `1' is generated. If the response is less than or equal to zero, a `0' is generated. The concatenated responses form a binary string that is converted into a numeric decimal value (the BSIF code). For example, if the \textbf{n} binary responses were $\{1, 0, 0, 1, 1\}$, the resulting decimal value would be `19'. Therefore, given \textbf{n} filters, the BSIF response will range between $0$ and $2^n - 1$.

\begin{figure*} % Tesselated Ocular Images (ocular, iris-only, iris-excluded)
    \centering
    \begin{subfigure}{.3\textwidth}
        \centering
        \includegraphics[width=0.5\textwidth]{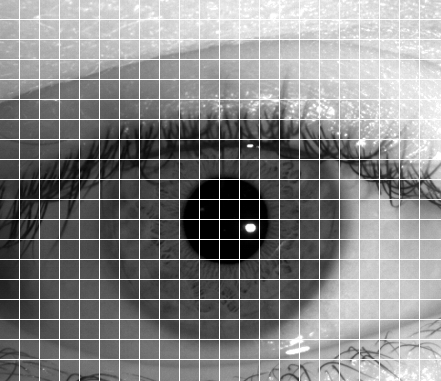}
        \caption{Ocular Image}
    \end{subfigure}
    \begin{subfigure}{.3\textwidth}
        \centering
        \includegraphics[width=0.4\textwidth]{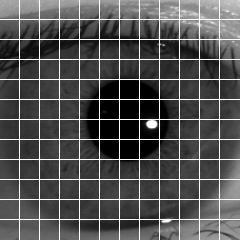}
        \caption{Iris-Only Image}
    \end{subfigure}
        \begin{subfigure}{.3\textwidth}
        \centering
        \includegraphics[width=0.5\textwidth]{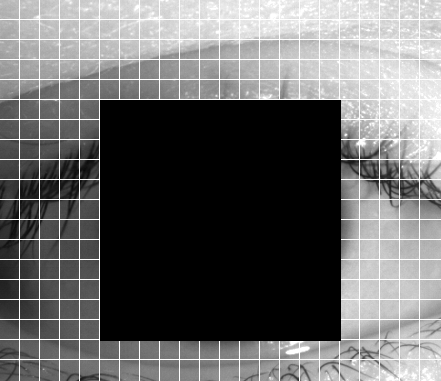}
        \caption{Iris-Excluded Ocular Image}
    \end{subfigure}
    \caption{Tessellations applied to the three image regions. The images are from \cite{doyle2013variation}.}
    \label{fig:tesselatedImages}
\end{figure*}

% removed this figure  (see Figure \ref{processFlow})
Our proposed method applies the texture descriptor to each of the NIR ocular images which were then tesselated into $20 \times 20$ pixel regions (see Figure~\ref{fig:tesselatedImages} for a visual representation). This tessellation was done in order to ensure the spatial information is included in the feature vector that is being created.   Histograms were generated for each of the tessellations, normalized,  and concatenated into a single feature vector. In order to provide consistent spatial information across each image, a geometric alignment was applied to the original NIR ocular image. The parameters chosen for this geometric alignment are similar to those proposed by~\cite{bobeldyk2016iris} and discussed in Section \ref{bioCOP2009Dataset} as well as shown in Figure~\ref{geometricAlignmentExample}.

\section{Datasets}
\label{datasetsSection}
Three separate datasets were used to conduct the experiments in this paper. The largest of the 3 datasets is the BioCOP2009 dataset. It is described in Section \ref{bioCOP2009Dataset}. Two other datasets were used for cross testing in order to demonstrate the generalizability of the proposed method. Those datasets are the Cosmetic Contact dataset (see Section \ref{cosmeticContactDataset}) and the GFI dataset (see Section \ref{gfiDataset}). It is also important to note that the two datasets used for cross testing were collected at an entirely different location than the BioCOP2009 dataset. The BioCOP2009 dataset was collected at West Virginia University, while the Cosmetic Contact and GFI dataset were both collected at Notre Dame University. Cross testing on datasets collected at different locations greatly decreases the chance that they will contain the same subjects while introducing substantial variability in the images due to changes in factors such as lighting and sensors.

\subsection{BioCOP2009 Dataset}
\label{bioCOP2009Dataset}
\begin{table*}
%\caption{The subset of the BioCOP2009 iris dataset that was used in the experiments}
\caption{Statistics of the BioCOP2009 dataset. The first column denotes the number of images that were initially present in the BioCOP2009 dataset. The second column lists the number of images that were successfully preprocessed by the COTS SDK in order to find the coordinates of the iris center and the iris radius. The third column presents the number of images that contained sufficient border pixels after the geometric alignment step.}
\label{bioCOP09ImageNumbers}
\centering

\begin{tabular*}{\textwidth}{c @{\extracolsep{\fill}} c c c c}  % This keeps the table inside the page width (not exceeding the borders)
\hline\hline
 \multirow{2}{*}{\textbf{Sensor}} & \textbf{Initial Number} & \textbf{Post SDK} & \textbf{Post Geometric}  \\
  &\textbf{of Images}&\textbf{Preprocessing}&\textbf{Alignment}\\
\hline
LG ICAM 4000 & 21,940 & 21,912 & 21,871 \\
CrossMatch I SCAN 2& 10,890 & 10,643& 9,121\\
Aoptix Insight & 10,980 & 10,979& 10,838\\
\hline
Total & 43,810 &43,534 & 41,830\\
\hline
%}
\end{tabular*}
\end{table*}

\begin{table*}
\caption{Statistics of the post processed BioCOP2009 dataset}
\label{sensorImageBreakdown}
\centering
\begin{tabular}{c c c c c}
\hline\hline
 \multirow{2}{*}{\textbf{Sensor}} &\textbf{Subjects with} & \textbf{Number of} & \textbf{Subjects with} & \textbf{Number of} \\
  & \textbf{Left Images}&\textbf{Left Images}&\textbf{Right Images}&\textbf{Right Images}\\
\hline
Aoptix&1096 & 5449 &1094 &5389  \\
CrossMatch&990 &4528&993&4593  \\
LG ICAM &1096 & 10,940& 1096&10,931 \\
\hline
Overall &1096&20,917 &1096&20,913  \\
\hline\hline
\end{tabular}
\end{table*}

\begin{figure*} % Example of geometrically aligned images
    \centering
    \begin{subfigure}{.3\textwidth}
        \centering
        \includegraphics[width=.75\textwidth]{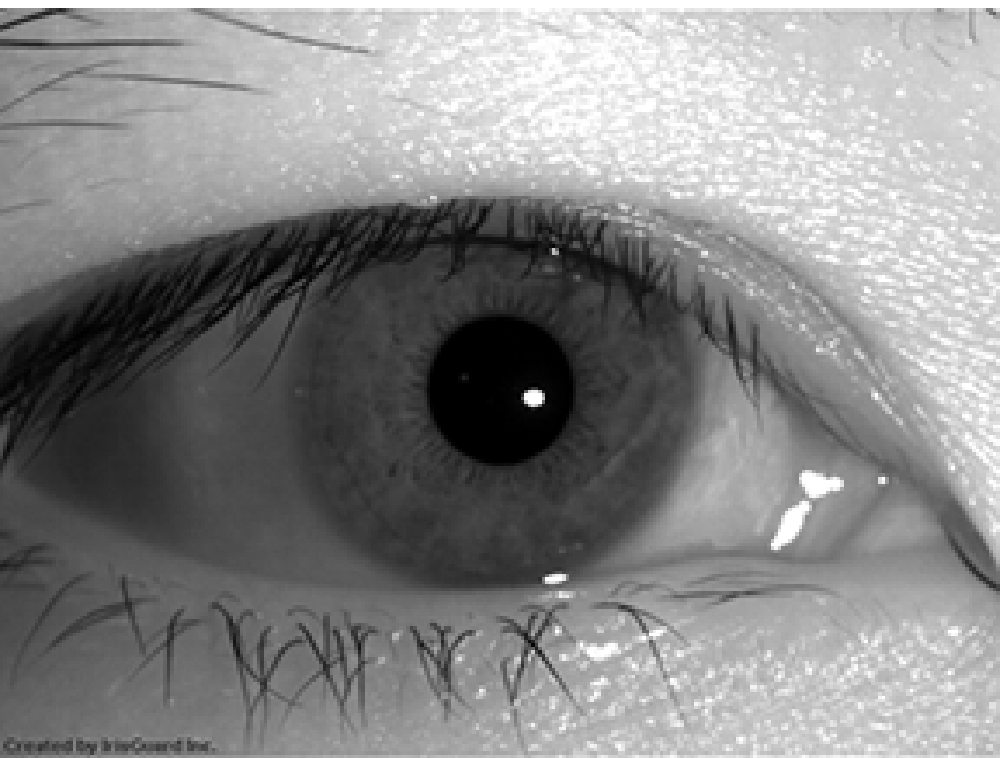}
        \caption{Before}
    \end{subfigure}
    \begin{subfigure}{.3\textwidth}
        \centering
        \includegraphics[width=.75\textwidth]{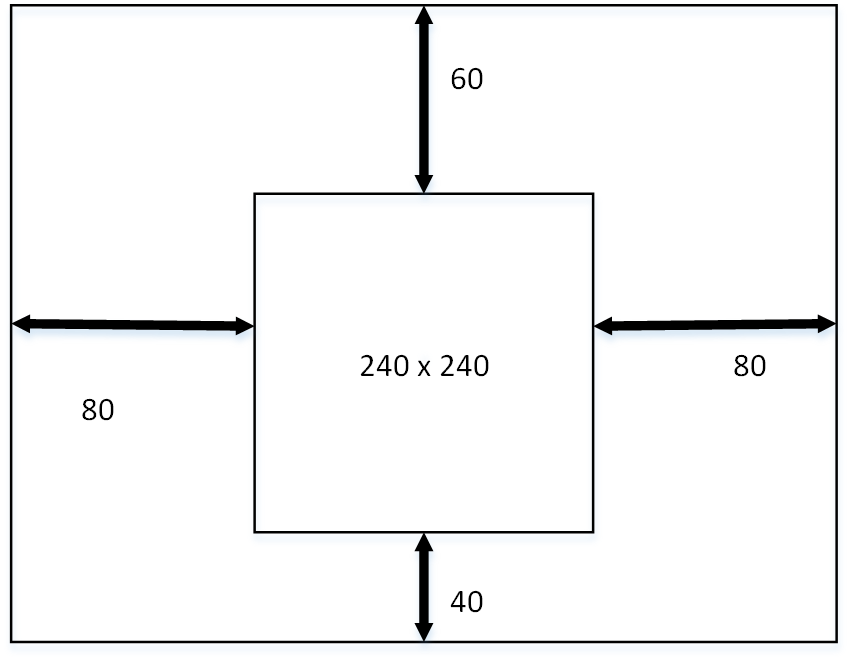}
        \caption{Geometric Alignment }
    \end{subfigure}
    \begin{subfigure}{.3\textwidth}
        \centering
        \includegraphics[width=.65\textwidth]{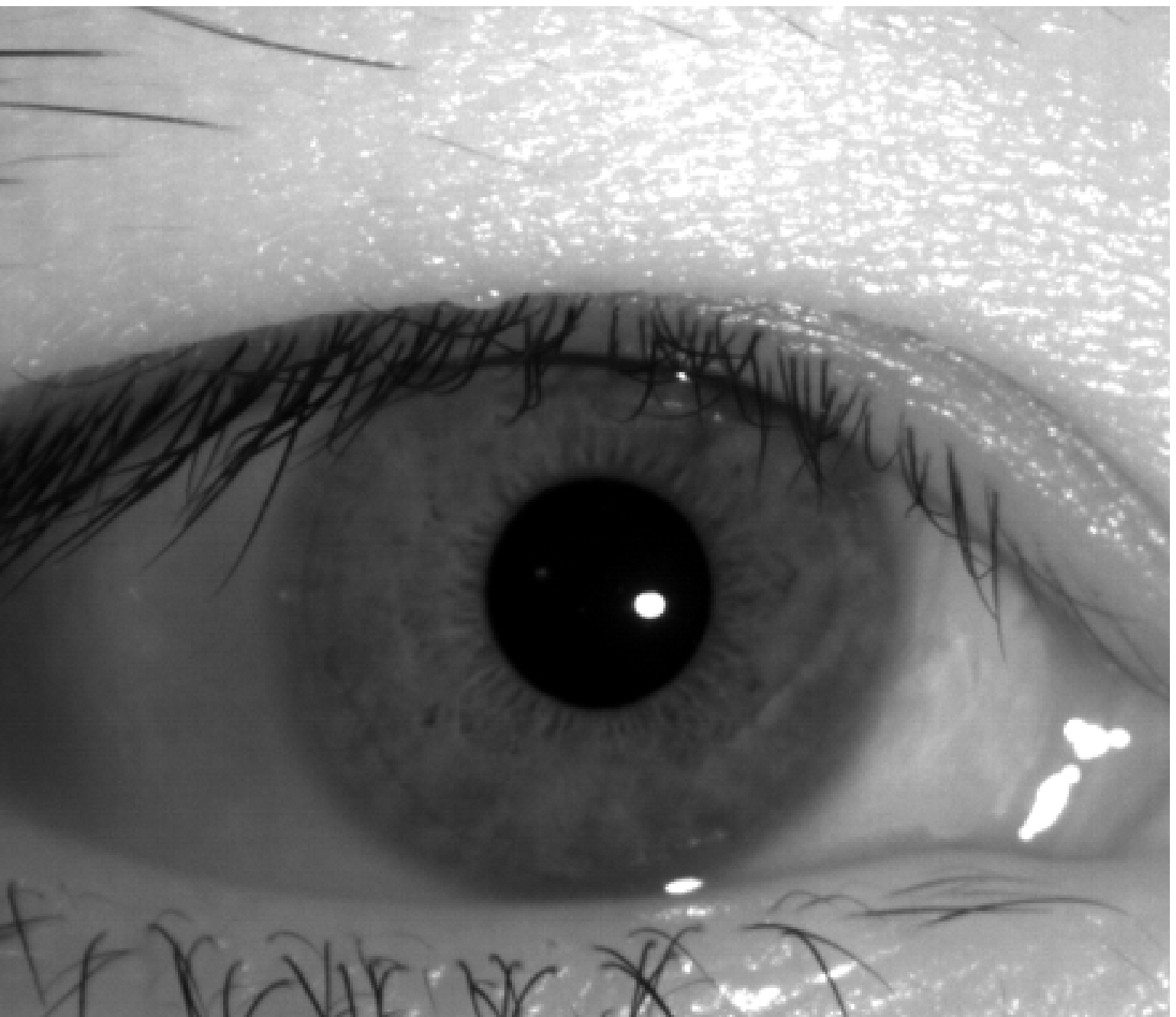}
        \caption{After}
    \end{subfigure}
    \caption{Example of a geometrically adjusted image. The image in (a) is from \cite{doyle2013variation}.}
    \label{geometricAlignmentExample}
\end{figure*}

The BioCOP2009 dataset contains 43,810 NIR ocular images captured with 3 different sensors: the LG ICAM 4000, CrossMatch I SCAN2 and Aoptix Insight. The LG and Aoptix sensors captured NIR ocular images of size $640\times 480$, while the CrossMatch sensor produced images of size  $480 \times 480$. Using a commercially available SDK, the images were preprocessed to find the coordinates of the iris center and the radius of the iris. During the preprocessing stage, 276 images were rejected as the software was unable to automatically locate those coordinates. In order to ensure that all images are spatially aligned, the images were geometrically adjusted using the method outlined in \cite{bobeldyk2016iris}. The geometric alignment centers the image, using the coordinates computed by the commercial SDK, and rescales the image, using bicubic interpolation, to a fixed size. Given that the CrossMatch sensor images were smaller than those in \cite{bobeldyk2016iris}, all of the images were aligned to the smaller dimension size of $400\times340$ (as opposed to $440\times380$ in \cite{bobeldyk2016iris}). A diagram displaying the pixel measurements, as well as a sample geometrically aligned image, are shown in Figure \ref{geometricAlignmentExample}. Images that did not contain sufficient border size after the geometric alignment were not used in the experiments (see Table \ref{bioCOP09ImageNumbers}). The majority (89\%) of those images were from the CrossMatch sensor, of which the initial image size was smaller ($480\times480$) than the other two sensors ($640\times480$); thereby making it more difficult to achieve the fixed border size.

There are 1096 total subjects in the post processed BioCOP2009 dataset, for a total of 41,830 images.  The Aoptix and LG ICAM sensors have a left eye image for every subject, while the CrossMatch sensor has 106 subjects with no left eye images. For the right eye, the LG ICAM has an image for every subject, the Aoptix has 2 subjects with no images and the CrossMatch has 103 subjects with no images. A summary of the sensor breakdown is shown in Table \ref{sensorImageBreakdown}. Throughout the rest of the document, the post processed BioCOP2009 dataset will simply be referred to as the `BioCOP2009 dataset'.

Five random subject-disjoint partitions of the BioCOP2009 dataset were created for both race and gender experiments. In each of those experiments, all of the images of a subject were used for either training or testing. Given some subjects have more images than others, the total number of training and testing images can fluctuate across the 5 random partitions. It is also important to note that the training and testing partitions contain images from all 3 sensors.

\subsection{Cosmetic Contact Dataset}
\label{cosmeticContactDataset}
In order to perform cross dataset testing, we used the Cosmetic Contact Lens dataset assembled by researchers at Notre Dame \cite{doyle2013variation}. The Cosmetic Contact Lens dataset contains images that are labeled with both race and gender labels. The dataset contains images collected by 2 separate sensors, the LG4000 and the AD100. For the LG4000 sensor, 3000 images were collected for training a classifier and 1200 images were collected for testing that classifier. For the AD100 sensor, 600 images were collected for training a classifier and 300 images were collected for testing that classifier. For the purposes of our experiments we only used the LG4000 sensor images. The rest of the paper will refer to the 3000 images collected from the LG4000 sensor as Cosmetic Contact Dataset One (CCD1)  and the 1200 verification images as Cosmetic Contact Dataset Two (CCD2).

The geometric alignment process that was used for the BioCOP2009 dataset (see Figure~\ref{geometricAlignmentExample}) was applied to the CCD1 and CCD2 datasets. After the geometric alignment procedure, only 4 images from CCD2 were discarded due to insufficient border size and no images were discarded from CCD1. During cross dataset testing, these 2 datasets were tested using the 5 SVM classifiers that were obtained from the 5 random partitions of the BioCOP2009 training set. Using the same SVM classifiers allows for a fair comparison between the prediction accuracies of the intra-dataset and cross-dataset test scenarios.

\subsection{GFI Dataset}
\label{gfiDataset}
The GFI dataset is a publicly available dataset that was assembled by researchers at Notre Dame University. It contains 3000 NIR ocular images, 1500 of which are from male subjects and 1500 from female subjects. There are 750 right and 750 left images for each of the aforementioned categories. The dataset was first used in \cite{tapia2014gender} but was discovered to contain multiple images from the same subjects (`an average of about six images per subject' \cite{tapia2016gender}). The dataset was corrected and used again in \cite{tapia2016gender} where it was stated to contain images from 1500 \textit{unique} subjects, 750 males and 750 females. The images were captured with a LG 4000 sensor \cite{tapia2016gender} and are labeled with the gender of the subject.

An additional GFI validation dataset was also available (also collected by Notre Dame) containing 3 images per eye of 324 subjects for a total of 972 left and 972 right NIR ocular images~\cite{tapia2016gender}.

\section{Experiments}
\label{experimentsSection}

\begin{table*}

\caption{Performance of the proposed race prediction method on the BioCOP2009 dataset: BSIF 8-bit 9x9 filter size,  LBP, LPQ}
\label{raceTextureDescriptorComparison}

\centering
\begin{tabular}{c l l l l}
\hline\hline
\multicolumn{5}{c}{\textbf{Race}}\\
\hline\hline
\textbf{Eye}&\textbf{Region}&\textbf{BSIF}  & \textbf{LBP} &\textbf{LPQ}\\
\hline
\multirow{3}{*}{Left}&Iris-Only&\color{red}$88.9\pm1.4$ &$86.5\pm1.5$ &$86.9\pm1.4$\\
&Iris-Excluded&$82.6\pm1.5$&\color{red}$88.0\pm1.5$&$79.6\pm0.9$\\
&Extended Ocular&\color{red}$89.8\pm 1.5$&$88.4\pm1.7$&$87.6\pm1.3$\\
\hline
\multirow{3}{*}{Right}&Iris-Only&\color{red}$88.6\pm1.2$&$85.9\pm0.8$&$87.1\pm0.8$\\
&Iris-Excluded&$82.7\pm0.6$             &\color{red}$85.5\pm0.8$       &$79.2\pm0.6$\\
&Extended Ocular&\color{red}$88.9\pm 1.1$&$87.1\pm0.9$&$87.5\pm0.8$\\
\hline
\end{tabular}
\end{table*}

\begin{table*}
\caption{Performance of the proposed gender prediction method on the BioCOP2009 dataset: BSIF 8-bit 9x9 filter size,  LBP, LPQ}
%\caption{BioCOP2009 Gender Texture Descriptor Comparison: BSIF 8-bit 9x9 filter size,  LBP, LPQ}
\label{genderTextureDescriptorComparison}

\centering
% The following command stretches the box to the textwidth
%\resizebox{\textwidth}{!}{\begin{tabular}{c c c c c}
%\begin{tabular*}{\textwidth}{c @{\extracolsep{\fill}}c c c c c}
\begin{tabular}{c l l l l}
\hline\hline
% \textbf{Left } & \textbf{Iris-Only}  & \textbf{Iris-Excluded}\\
% \textbf{or Right} & \textbf{Accuracy} & \textbf{Ocular Accuracy}\\
\multicolumn{5}{c}{\textbf{Gender}}\\
\hline\hline
\textbf{Eye}&\textbf{Region}&\textbf{BSIF}  & \textbf{LBP} &\textbf{LPQ}\\
\hline
\multirow{3}{*}{Left}&Iris-Only&\color{red}$78.9\pm1.0$    &\color{red}$78.9\pm0.2$        &$74.9\pm1.0$\\
&Iris-Excluded      &\color{red}$82.2\pm1.4$                               &\color{red}$82.9\pm0.9$        &\color{red}$81.8\pm0.8$\\
&Extended Ocular    &\color{red}$85.9 \pm 0.7$      &$84.1\pm0.5$                               &$82.4\pm0.8$\\
\hline
\multirow{3}{*}{Right}&Iris-Only&\color{red}$79.2\pm0.8$       &\color{red}$79.8\pm1.1$           &$74.9\pm1.2$\\
&Iris-Excluded      &\color{red}$82.1\pm0.9$                                     &\color{red}$82.0\pm0.6$                             &$80.6\pm1.4$\\
&Extended Ocular    &\color{red}$85.2\pm1.1$        &$84.0\pm0.7 $       &$81.4\pm1.2$\\
\hline
\end{tabular}
\end{table*}

\subsection{Race}
\label{raceSubsection}

Of the 1096 subjects contained in the BioCOP2009 dataset, 849 of them were labeled as `Caucasian', with the remaining 247 labeled with a variety of other classes (i.e., Asian, Hispanic, African). In order to create an equal number of subjects in each of the classes, 247 of the 849 Caucasian subjects were randomly selected. The remaining 602 Caucasian subjects were not used in the race prediction experiments. 60\% of the subjects were randomly selected to be in the training set while the remaining 40\% were selected for the test set, resulting in 148 subjects for training and 99 subjects for testing. This random selection was repeated 5 times resulting in 5 subject-disjoint training and testing sets. An SVM classifier was trained using the images from the 148 subjects selected for the training partition. Images from all 3 sensors were used during the training and testing stages. Over the 5 random iterations, there were $5656\pm34$ images in the training dataset and $3749 \pm 34$ images in the test set.\footnote{Some subjects may have more images than others}

\subsubsection{BioCOP2009 Race Results}

\begin{figure*}
\center
\includegraphics[width=0.7\textwidth]{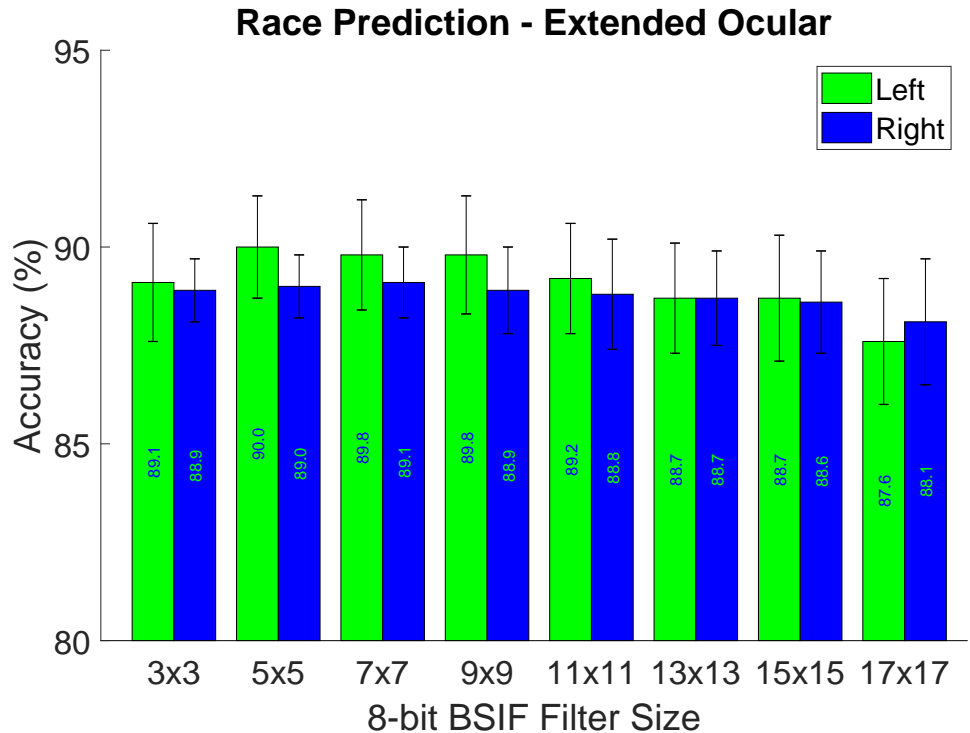}
\caption{Race prediction results using the extended ocular region (BioCOP2009 using 8-bit BSIF).}
\label{figEightBitRace}
\end{figure*}

The 8-bit BSIF was used in this work as a compromise between prediction accuracy and computational processing time. While 9-bit or 10-bit BSIF may provide slightly better results, the increased requirement of memory and processing time to perform each experiment was quite substantial given the large size of the BioCOP2009 dataset. An SVM classifier was trained on each of the 5 training sets using the extracted BSIF features described in Section \ref{featureExtractionSection}. The test data was classified using each respective SVM model. The resulting prediction accuracy using filter sizes in the range of $3\times3$ to $17\times17$ is shown in Figure \ref{figEightBitRace}. The prediction accuracy varies slightly across each of the different filter sizes; however, there is no significant difference in performance across the filter sizes.

\begin{table*}
\centering
\caption{Race prediction confusion matrix for the extended ocular region (BioCOP2009 using 8-bit BSIF with a 9x9 filter).}
\begin{tabular}{c c c |c c}
\hline\hline
&\multicolumn{2}{c}{\textbf{Left}}&\multicolumn{2}{c}{\textbf{Right}}\\
\hline
&\textbf{Predicted}&\textbf{Predicted}&\textbf{Predicted}&\textbf{Predicted}\\
&\textbf{Caucasian}&\textbf{Non-Caucasian}&\textbf{Caucasian}&\textbf{Non-Caucasian}\\\hline
\textbf{Actual Caucasian}&$91.7\%1.3\pm\%$&$8.3\%\pm1.3\%$&$90.5\%\pm1.9\%$&$9.5\%\pm1.9\%$\\
\textbf{Actual Non-Caucasian}&$12.1\%\pm2.8\%$&$87.9\%\pm2.8\%$&$12.8\%\pm2.7\%$&$87.2\%\pm2.7\%$ \\
%\textbf{Actual Brown}&$3501\pm33.9$&$509.6\pm58.1$   \\
%\textbf{Actual Not Brown}&$439.2\pm32.0$&$5135.8\pm54.8$ \\
\hline\hline
\end{tabular}
\label{raceOcularConfusionMatrix}
\end{table*}

\begin{table*}
\centering
\caption{Gender prediction confusion matrix for the extended ocular region (BioCOP2009 using 8-bit BSIF with a 9x9 filter).}
\begin{tabular}{c c c |c c}
\hline\hline
&\multicolumn{2}{c}{\textbf{Left}}&\multicolumn{2}{c}{\textbf{Right}}\\
\hline
&\textbf{Predicted}&\textbf{Predicted}&\textbf{Predicted}&\textbf{Predicted}\\
&\textbf{Female}&\textbf{Male}&\textbf{Female}&\textbf{Male}\\\hline
\textbf{Actual Female}&$83.2\%\pm1.8\%$&$16.8\%\pm1.8\%$&$82.2\%\pm2.2\%$&$17.8\%\pm2.2\%$\\
\textbf{Actual Male}&$11.4\%\pm1.7\%$&$88.6\%\pm1.7\%$&$11.8\%\pm1.8\%$&$88.2\%\pm1.8\%$ \\
%\textbf{Actual Brown}&$3501\pm33.9$&$509.6\pm58.1$   \\
%\textbf{Actual Not Brown}&$439.2\pm32.0$&$5135.8\pm54.8$ \\
\hline\hline
\end{tabular}
\label{genderOcularConfusionMatrix}
\end{table*}

\subsubsection{Iris-excluded Ocular Region vs. Iris-Only Region }
Previous work in this field \cite{tapia2014gender,tapia2016gender} has predominantly focused on the iris-only portion of the captured NIR ocular images. Bobeldyk and Ross \cite{bobeldyk2016iris} showed, for gender prediction using BSIF, that the ocular region provides greater gender prediction accuracy than the iris-only region. We have performed a similar experiment in order to test the prediction accuracy of the iris-only and iris-excluded ocular image regions (see Figure~\ref{irisRegionExamples}). The results of these experiments are shown in Table~\ref{irisExcludedIrisOnlyRace}. For race, the iris-only region provides a \textit{greater} prediction accuracy using BSIF than the iris-excluded ocular region, while the opposite is true for gender prediction.

\begin{table*} %\label irisExcludedIrisOnlyEthnicity
\caption{Race prediction using the iris-excluded, iris-only and extended ocular regions (BioCOP2009 using 8-bit BSIF with a 9x9 filter).}
\label{irisExcludedIrisOnlyRace}

\centering
% The following command stretches the box to the textwidth
%\resizebox{\textwidth}{!}{\begin{tabular}{c c c c c}
%\begin{tabular*}{\textwidth}{c @{\extracolsep{\fill}}c c c c c}
\begin{tabular}{c c c c}
\hline\hline
% \textbf{Left } & \textbf{Iris-Only}  & \textbf{Iris-Excluded}\\
% \textbf{or Right} & \textbf{Accuracy} & \textbf{Ocular Accuracy}\\
\multicolumn{4}{c}{\textbf{Race Prediction Accuracy ($\%$) }}\\
\hline\hline
\multirow{2}{*}{\textbf{Eye}} & \textbf{Iris-Only}  & \textbf{Iris-Excluded} & \textbf{Extended}\\
 & \textbf{Accuracy} & \textbf{Ocular Accuracy}&\textbf{Ocular Accuracy}\\
\hline
Left & $88.9 \pm 1.4$  & $82.6  \pm 1.5$ & $89.8\pm1.5$ \\
Right & $88.6\pm 1.2$  & $82.7 \pm 0.6$ & $89.9\pm1.1$ \\
\hline
%\end{tabular}}
\end{tabular}
\end{table*}

\subsubsection{Cross Dataset Testing}
\label{raceCrossDatasetSection}
\begin{figure*}[t!]
\centering
\begin{subfigure}[t]{0.24\textwidth}
\centering
\includegraphics[height=0.8in]{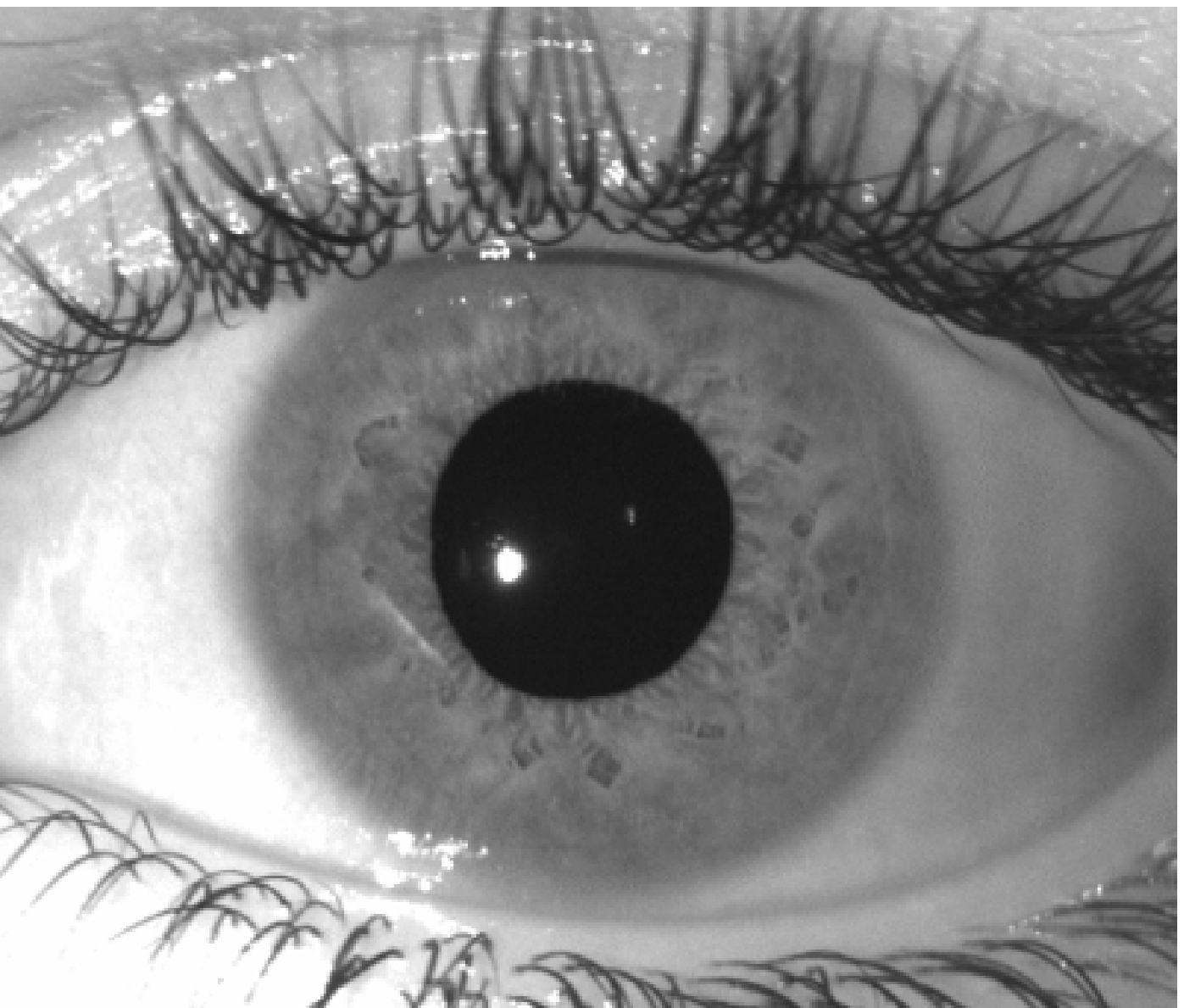}
\caption{Non-Caucasian}
\end{subfigure}
\begin{subfigure}[t]{0.24\textwidth}
\centering
\includegraphics[height=0.8in]{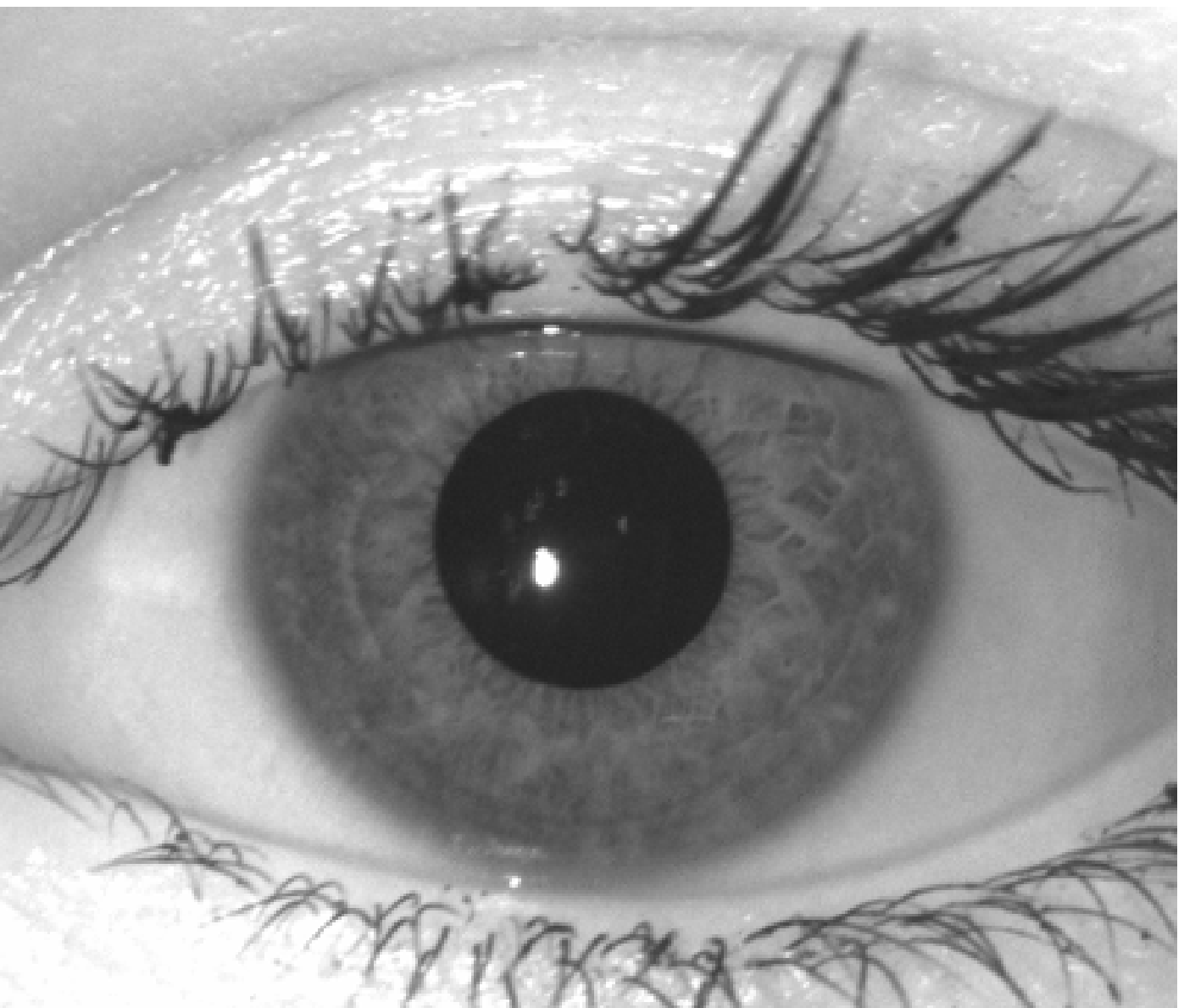}
\caption{Non-Caucasian}
\end{subfigure}
\begin{subfigure}[t]{0.24\textwidth}
\centering
\includegraphics[height=0.8in]{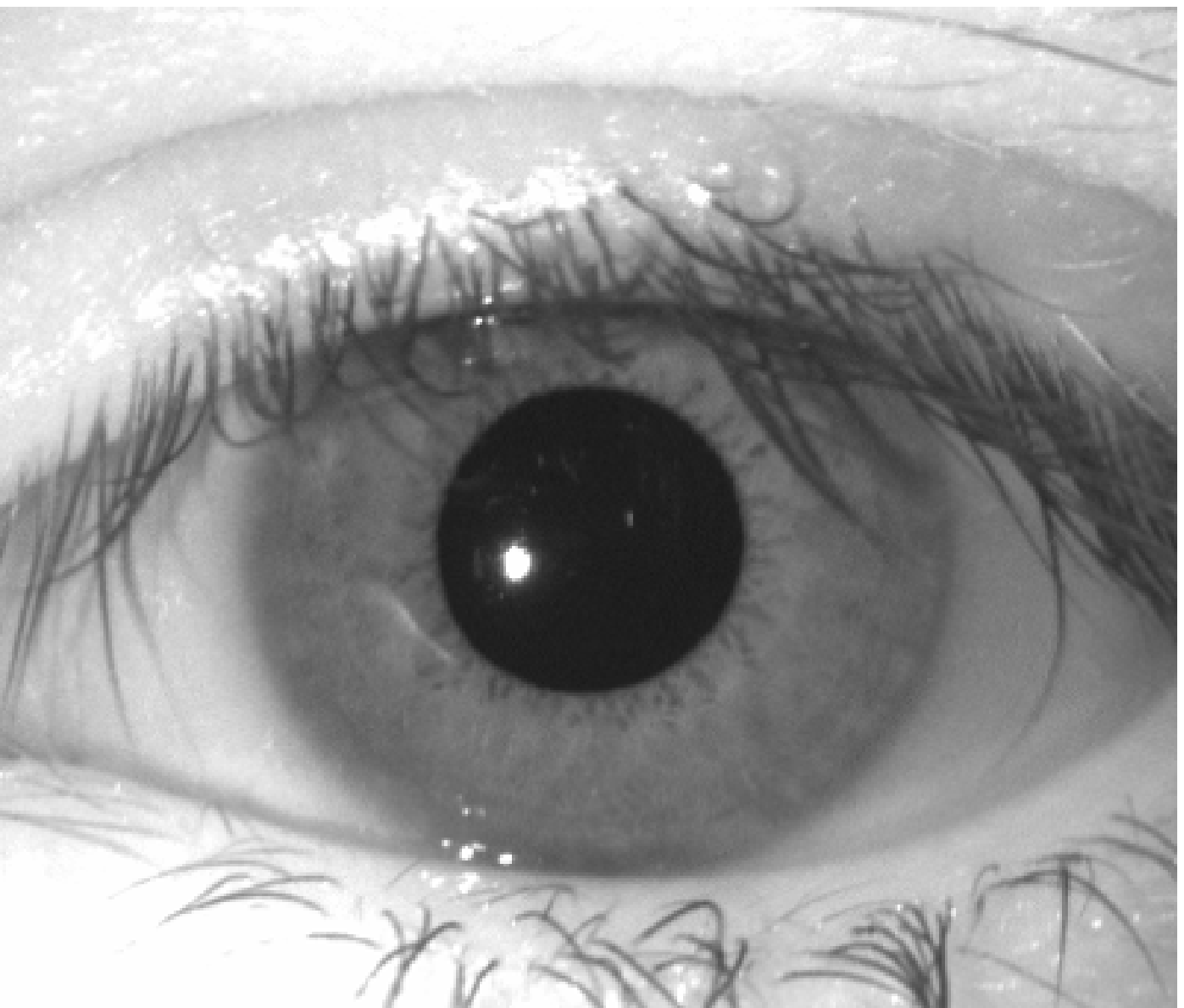}
\caption{Caucasian}
\end{subfigure}
\begin{subfigure}[t]{0.24\textwidth}
\centering
\includegraphics[height=0.8in]{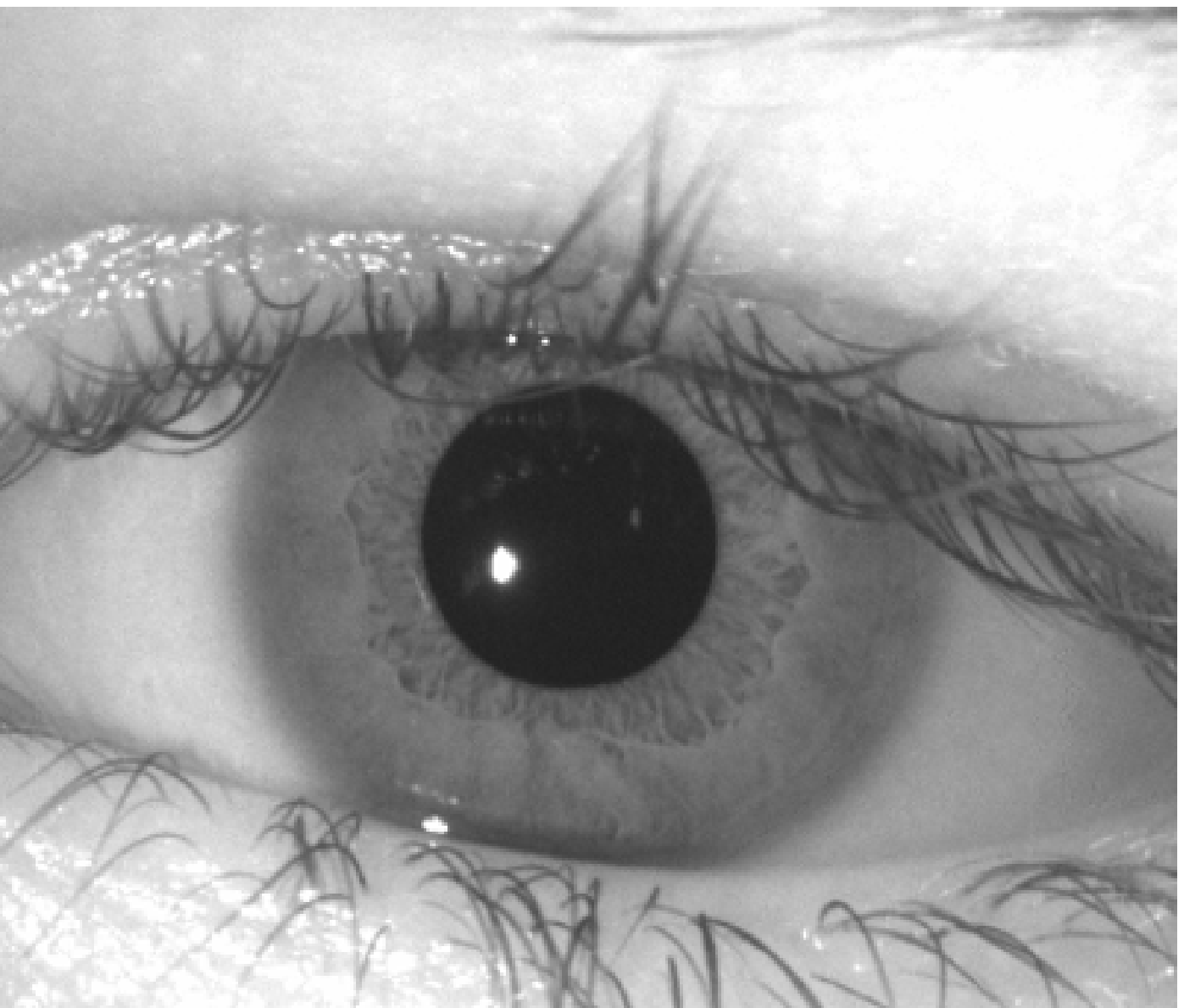}
\caption{Caucasian}
\end{subfigure}
\caption{Misclassified images: (a) and (b) were classified as Caucasian, (c) and (d) were classified as Non-Caucasian. The images are from \cite{doyle2013variation}.}
\label{misclassifiedRaceImages}
\end{figure*}

It is not uncommon for a method to perform well when training and testing is conducted using the same dataset. In order to demonstrate the generalizability of the proposed algorithm, we trained on the BioCOP2009 dataset and tested on the CCD1 and CCD2 datasets described earlier. The 5 trained SVM models that were generated using the BioCOP2009 dataset were used to classify the images in CCD1 and CCD2. It should be noted that subjects from the BioCOP2009 dataset were labeled as `Caucasian' while those in the CCD1 and CCD2 datasets were labeled as `White'.  Both the CCD1 and CCD2 datasets contain images of people with contacts, without contacts and with cosmetic contacts. CCD1 contains 500 left and 500 right eye images with no contacts, CCD2 contains 200 left and 200 right eye images with no contacts. Only the images without contacts were used in our experiments. The results are shown in Table \ref{raceCrossDatasetTesting}. The resulting prediction accuracy of the cross dataset experiments is comparable to that of the same dataset experiments which supports the hypothesis that the proposed approach does generalize. Some images that were misclassified are shown in Figure \ref{misclassifiedRaceImages}.

\begin{table*}
\caption{Race cross dataset testing (8-bit BSIF with a 9x9 filter).}
\label{raceCrossDatasetTesting}
\centering
\begin{tabular}{c c c c}
\hline\hline
\multicolumn{4}{c}{\textbf{Race}}\\
\hline\hline
\textbf{Training}&\textbf{Testing} & \textbf{Eye} & \textbf{Prediction Accuracy (\%)}\\
\hline
\multirow{4}{*}{BioCOP2009}& \multirow{2}{*}{CCD1}&Left & $80.2 \pm 1.3$ \protect\footnotemark\\
 & &Right & $90.3\pm 1.7$ \\
& \multirow{2}{*}{CCD2}&Left &$87.3 \pm 4.5$ \\
 & &Right &  $90.8 \pm 1.6$ \\
\hline
\end{tabular}
\end{table*}
\footnotetext{The lower prediction accuracy of the left eye could be attributed to the non symmetric composition of the subject pool between left and right eye images (of subjects that are not wearing contacts). If the contact lens images are also included, the prediction accuracy increases to $88.9\%\pm1.2\%$}

\subsection{Gender}
\label{genderSubsection}
\subsubsection{BioCOP2009 Gender Results}
\label{genderBioCOP2009experiments}

\begin{figure*}
\center
\includegraphics[width=0.6\textwidth]{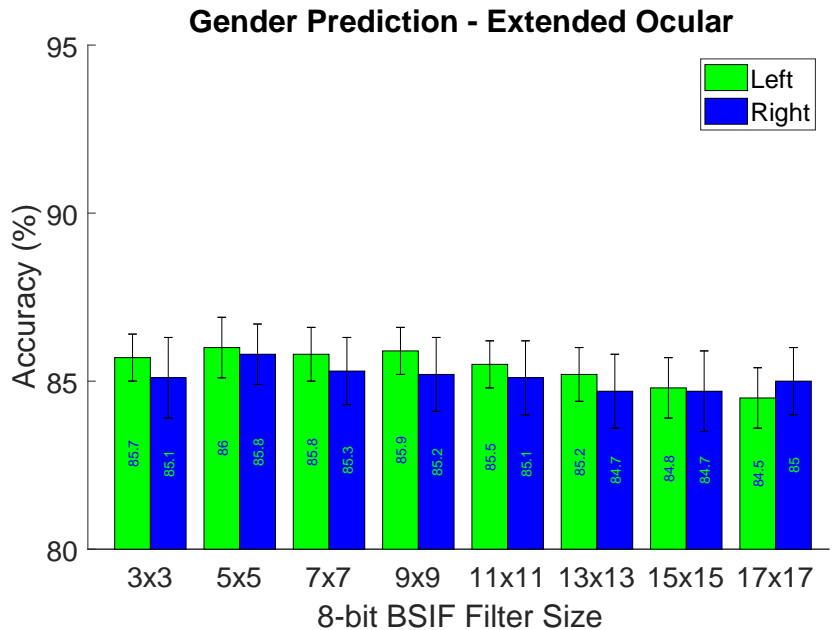}
\caption{Gender prediction results using the extended ocular region (BioCOP2009 using 8-bit BSIF).}
\label{fig:EightBitGender}
\end{figure*}

Of the 1096 subjects contained in the BioCOP2009 dataset, 467 are labeled male and 629 are labeled female.\footnote{It should be noted that societal and personal interpretation of gender may consider more than a simple `male' and `female' label. For example, at the time of this paper's publication, Facebook has 71 gender options.} In order to assign an equal number of subjects to each class, 467 of the 629 available female subjects were randomly selected. The remaining 162 female subjects were not used for these experiments. 60\% of the subjects were randomly chosen to be in the training set (280 subjects and their associated images) while the remaining 40\% were placed in the test set (187 subjects and their associated images). This process of random selection was repeated 5 times, creating 5 different subject-disjoint sets for training and testing. An SVM classifier was trained on images from the training set. Images from all 3 sensors were pooled together during the training and testing process. Over the 5 random iterations for the left eye there were $10,727 \pm 3.4$  images used for training and $7,156 \pm 3.4$ images used for testing.\footnote{Some subjects may have more images than others} For the right eye there were $10,720 \pm 11.3$ images used for training and $7,159 \pm 11.3$ used for testing.\footnote{Some subjects may have more images than others} The results of the experiments across all of the BSIF filter sizes are shown in Figure~\ref{fig:EightBitGender}. The prediction accuracy only varies slightly across each of the different filter sizes.

\subsubsection{Iris-excluded Ocular versus Iris-Only}
Previous literature has shown that the iris-excluded ocular region (as shown in Figure \ref{irisRegionExamples})  provides greater sex discrimination than the iris region when the BSIF descriptor is used ~\cite{bobeldyk2016iris}.  A separate feature vector was generated from each of the two regions: iris-excluded ocular and iris-only (see Figure \ref{irisRegionExamples}). The results are shown in Table \ref{genderIrisOnlyIrisExcludedResults} and confirm the results found in \cite{bobeldyk2016iris}.

\begin{table*} %\label{genderIrisOnlyIrisExcludedResults}
\caption{Gender prediction results using the iris-excluded, iris-only and extended ocular regions (BSIF 8bit-9x9 filter size)}
\label{genderIrisOnlyIrisExcludedResults}
\centering
\begin{tabular}{c c c c}
\hline\hline
\multicolumn{4}{c}{\textbf{Gender Prediction Accuracy ($\%$) }}\\
\hline\hline
\multirow{2}{*}{\textbf{Eye}} & \textbf{Iris-Only}  & \textbf{Iris-Excluded}& \textbf{Extended} \\
& \textbf{Accuracy} & \textbf{Ocular Accuracy}&\textbf{Ocular Accuracy} \\
\hline
Left&$78.9 \pm 1.0 $  & $82.2 \pm 1.3$ & $85.9\pm0.7$ \\
Right& $79.2 \pm 0.8 $  & $ 82.1 \pm 0.9 $ & $85.2\pm 1.1$ \\
\hline
\end{tabular}
\end{table*}

%
%\subsubsection{Multiclass SVM}

\subsubsection{Cross Dataset Testing}
\label{genderCrossDatasetSection}
In order to validate the proposed method and ensure generalizability of the algorithm to images originating from outside of the BioCOP2009 dataset, we chose to cross test on the following datasets: CCD1, CCD2, GFI, and GFI-validation. Each of these datasets were made available by the researchers at Notre Dame \cite{doyle2013variation}. It was important to choose a dataset originating from a separate location than where the BioCOP2009 dataset was collected\footnote{The BioCOP2009 dataset was collected at West Virginia University} in order to reduce the chance of the same identity being included in both of the datasets. The CCD1 and CCD2 datasets provide both gender and race labels for each of the images, while the GFI and GFI-validation datasets provides only gender. The CCD1 and CCD2 datasets contains images of subjects with contacts, without contacts and with cosmetic contacts. CCD1 contains 500 left and 500 right eye images with no contacts, CCD2 contains 200 left and 200 right eye images with no contacts. Only the images of subjects without contacts were used in the experiments.

\begin{figure*}[t!]   %\label{misclassifiedGenderImages}
\centering
\begin{subfigure}[t]{0.24\textwidth}
\centering
\includegraphics[height=0.8in]{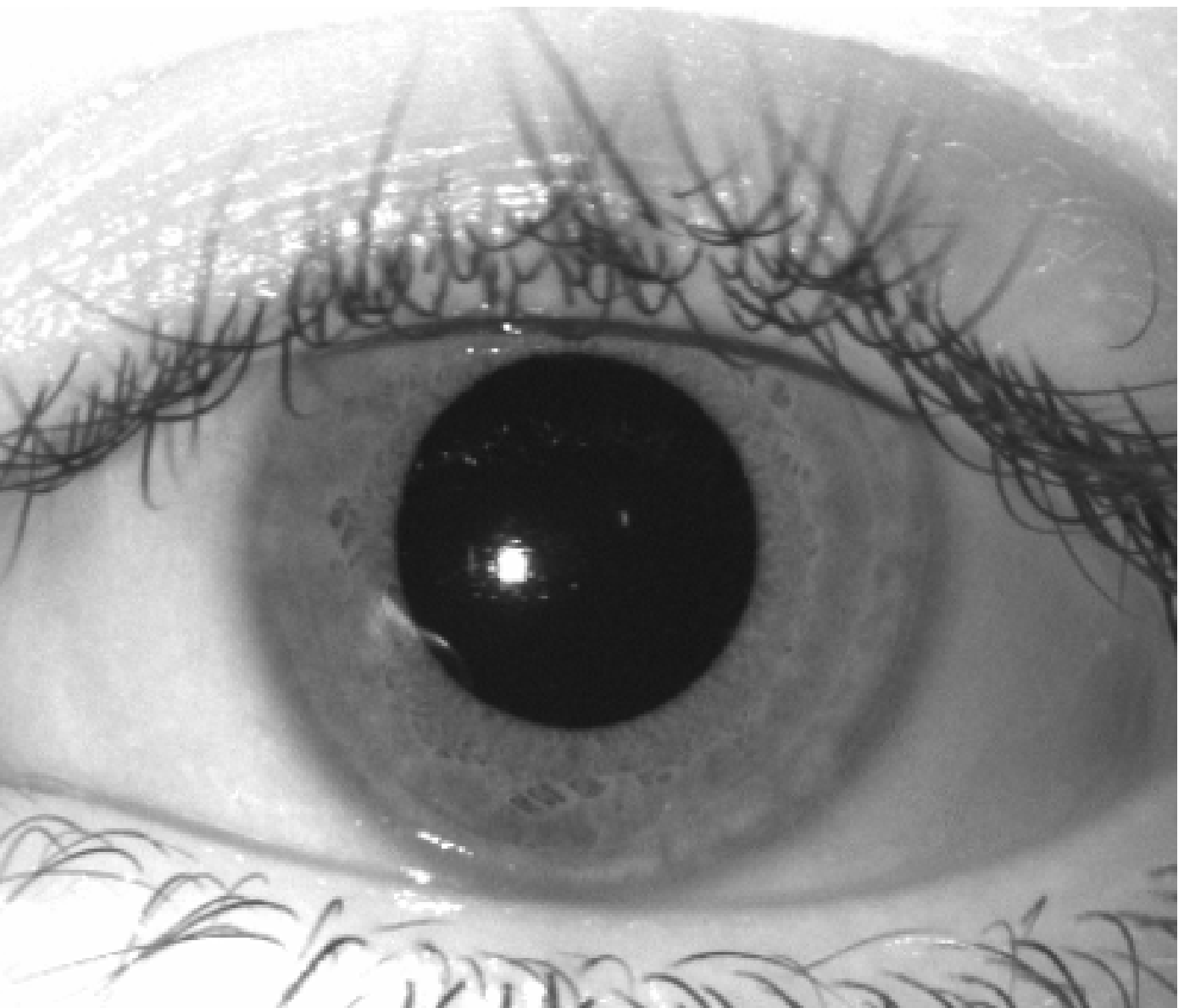}
\caption{Male}
\end{subfigure}
\begin{subfigure}[t]{0.24\textwidth}
\centering
\includegraphics[height=0.8in]{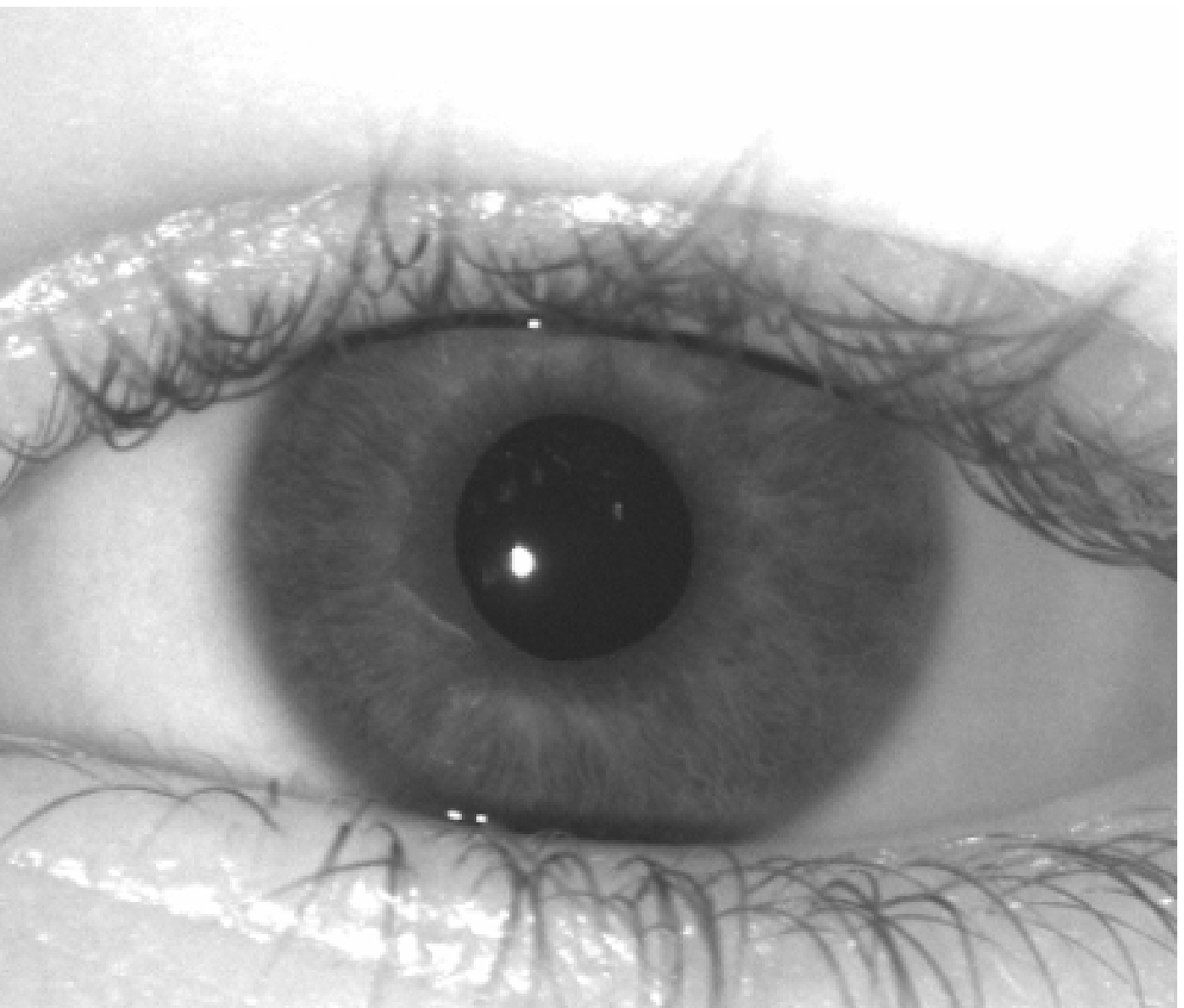}
\caption{Male}
\end{subfigure}
\begin{subfigure}[t]{0.24\textwidth}
\centering
\includegraphics[height=0.8in]{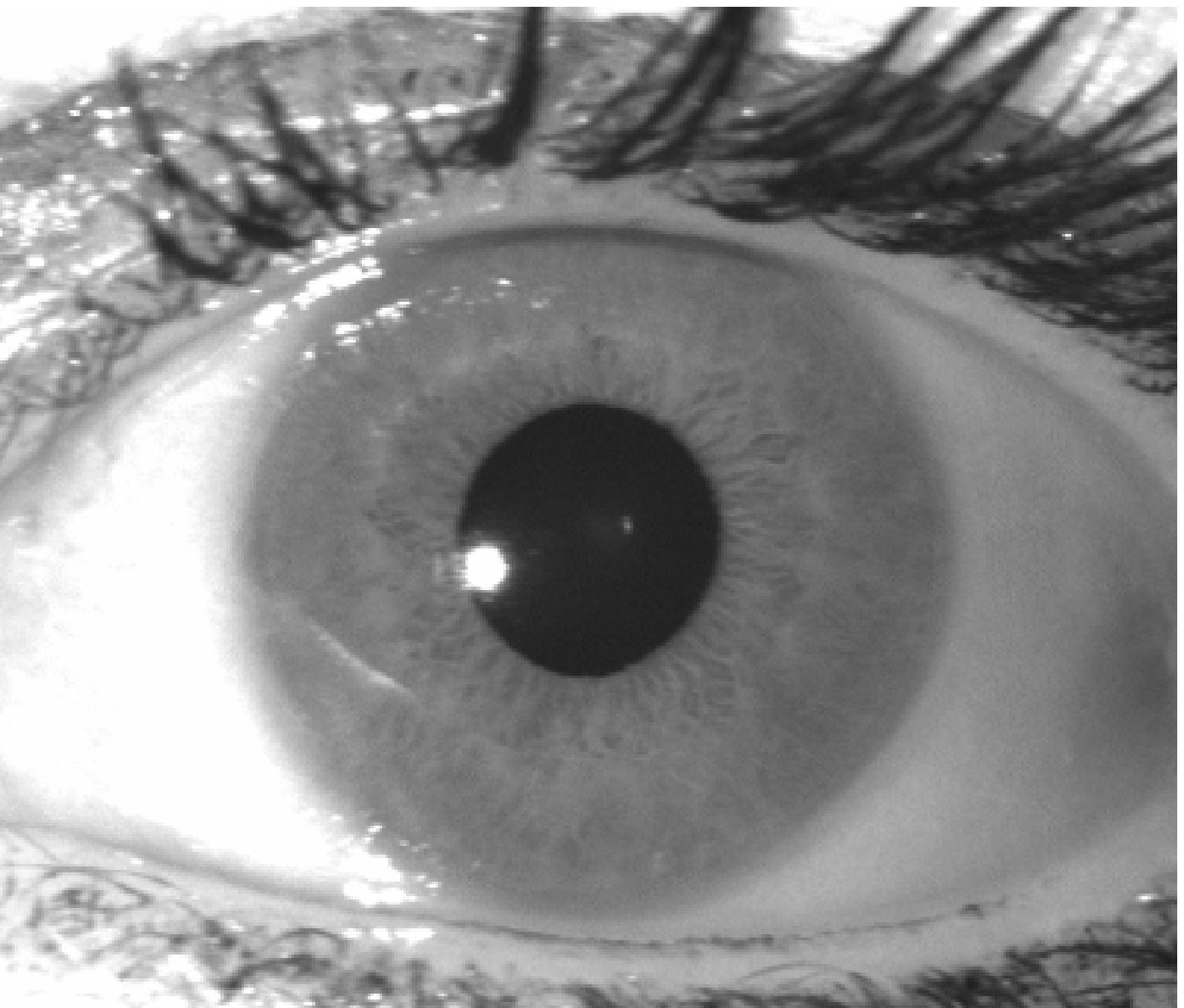}
\caption{Female}
\end{subfigure}
\begin{subfigure}[t]{0.24\textwidth}
\centering
\includegraphics[height=0.8in]{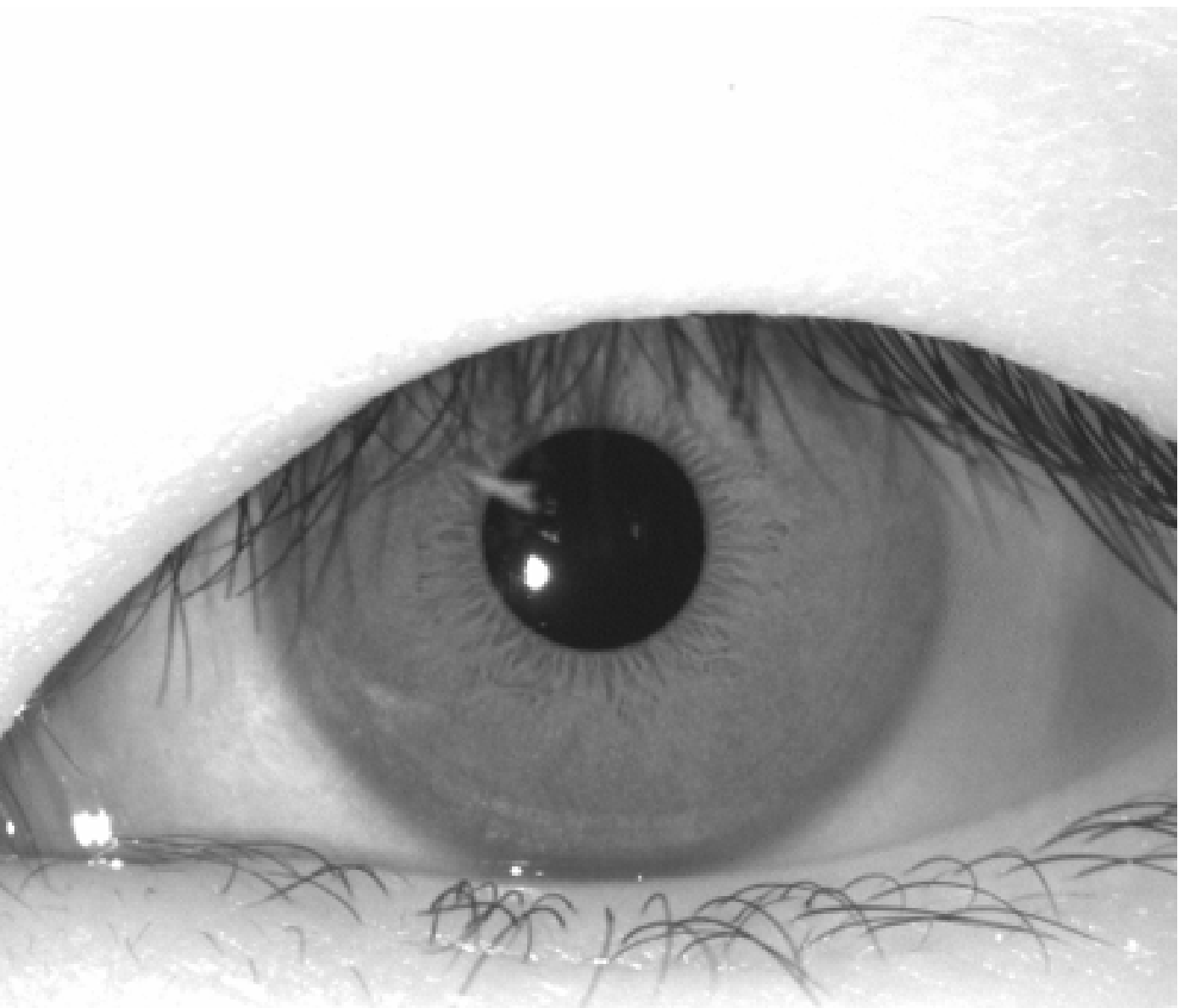}
\caption{Female}
\end{subfigure}
\caption{Misclassified images: (a) and (b) were classified as female, (c) and (d) were classified as male. The images are from \cite{doyle2013variation}.}
\label{misclassifiedGenderImages}
\end{figure*}

 The 5 trained SVM models that were generated using the BioCOP2009 dataset were used to classify images in each of the 4 selected datasets (CCD1, CCD2, GFI, GFI-validation). The results are shown in Table~\ref{genderCrossDatasetTesting}. The prediction accuracy for classification of images from CCD1 and CCD2 was about 10\% less than that of the GFI and GFI-validation datasets. We believe this may be due to the increased number of images per subject in the cosmetic contact dataset. The images in the GFI dataset, on the other hand, contain only 1 image per subject. Some images that were misclassified are shown in Figure \ref{misclassifiedGenderImages}.

%/user/bobeldy3/matlab/biocop09-3/journal/gender/crossTestCC09-3/predictGenderCCwith09dash3V3.m

\begin{table*}
\caption{Gender prediction results in a cross-dataset scenario where training and testing are done on different datasets (BSIF 8bit-9x9 filter size).}
\label{genderCrossDatasetTesting}
\centering
\begin{tabular}{c c c c}
\hline\hline
\multicolumn{4}{c}{\textbf{Gender}}\\
\hline\hline
\textbf{Training}&\textbf{Testing} & \textbf{Eye} & \textbf{Prediction Accuracy}\\
\hline
\multirow{8}{*}{BioCOP2009}& \multirow{2}{*}{CCD1}&Left & $ 75.3\pm2.1 $  \\
 & &Right & $76.8\pm 2.9$  \\
& \multirow{2}{*}{CCD2}&Left &$ 72.3\pm  4.0$\\
 & &Right &   $77.8\pm 4.1$ \\
 & \multirow{2}{*}{GFI}&Left &$84.4 \pm 0.8 $ \\
 & &Right &  $84.3\pm 0.5 $\\
 &GFI-&Left &$84.2 \pm1.2 $ \\
 &Validation &Right &  $82.6 \pm 1.3$ \\
\hline
\end{tabular}
\end{table*}

\subsection{Race and Gender}

The following two subsections will analyze the impact of race on gender prediction and the impact of gender on race prediction. It should be noted that regardless of which experiment was performed, there was very little variation in prediction accuracy between the left and right eye images.

\subsubsection{Gender prediction: Caucasian versus Non-Caucasian Analysis}
In order to determine if predicting gender is a more challenging problem for either Caucasians or Non-Caucasians, 4 additional experiments were performed: (a) training and testing on Caucasian subjects; (b) training on Caucasian subjects and testing on Non-Caucasian subjects; (c) training and testing on Non-Caucasian subjects; (d) training on Non-Caucasian subjects and testing on Caucasian subjects.

%performed. The first, will train and test on images from Caucasian subjects. The second, will train on images from Caucasian subjects and test on images from Non-Caucasian subjects. The third, will train and test on images from Non-Caucasian subjects. The fourth, will train on images from Non-Caucasian subjects and test on images from Caucasian subjects.
%
%The first two experiments will train and test on the same race class, while the second two will train on one class and test on the other class.

Training and testing on only the Caucasian class results in a $\sim$6\% increase in prediction accuracy when compared to training and testing on only the Non-Caucasian class. The decrease in prediction accuracy for the Non-Caucasian class could be attributed to the multiple race labels that were assigned to the Non-Caucasian class (see Section ~\ref{raceSubsection}). The results are shown in Table~\ref{genderCaucasianNonCaucasianContrast}.

Training on either race class and cross testing on the other race class results in an $\sim$80\% prediction accuracy. It can be observed that there is a slight increase in prediction accuracy when training on the Non-Caucasian class and testing on the Caucasian class ($\sim$1-2\%). The results are shown in Table~\ref{genderCaucasianNonCaucasianContrast}.

%It can also be observed, that unlike the intra-class testing, there is very little difference in the prediction accuracy ($\sim$1-2\%) regardless of which race class was trained/tested on.

\subsubsection{Race prediction: Male versus Female Analysis}

In order to determine if predicting race was a more challenging problem for either males or females, 4 additional experiments were conducted: (a) training and testing on male subjects; (b) training and testing on female subjects; (c) training on male subjects and testing on females; and (d) training on females and testing on males.

Training and testing on only male subjects results in a $\sim$3-5\% increase over training and testing on only female subjects.\footnote{These prediction results agree with the findings of the earlier work by Lagree and Bowyer \cite{lagree2011predicting}, who observed an increase in prediction accuracy when training and testing on male subjects, compared to female subjects.} There was a significant decrease in prediction accuracy when training on male subjects and testing on female subjects ($\sim$14\%). There was no decrease in prediction accuracy when training on female subjects and testing on male subjects. The absence or presence of makeup in the female images may make it more difficult for the male-only trained model to predict race from the female images, but additional research should be performed to fully explore the difference in prediction accuracies. The results are summarized in Table~\ref{raceMaleFemaleContrast}.

\begin{table*}
\caption{Gender prediction results for intra-race and inter-race training and testing (BioCOP2009 using 8-bit BSIF with a 9x9 filter).}
\label{genderCaucasianNonCaucasianContrast}
\centering
\begin{tabular}{c c c c}
\hline\hline
\multicolumn{4}{c}{\textbf{Gender}}\\
\hline\hline
 %\textbf{Train On} & \textbf{Test On} & \textbf{Right or Left} & \textbf{Prediction Accuracy} \protect\footnotemark\\
 \multirow{2}{*}{ \textbf{Train On}} &\multirow{2}{*}{ \textbf{Test On}} & \multirow{2}{*}{\textbf{Eye}} & \textbf{Prediction} \\
 &&&\textbf{Accuracy}\\
 \hline
 \multirow{4}{*}{Caucasian}&  \multirow{2}{*}{Caucasian} & Left & $87.9 \pm 1.3 $ \\
 &   &Right & $87.2 \pm 1.1 $ \\
 & \multirow{2}{*}{Non-Caucasian} &Left & $ 77.5$ \\
 & & Right & $ 78.5$\\
 \hline
 \multirow{4}{*}{Non-Caucasian} & \multirow{2}{*}{Non-Caucasian} &Right & $81.3 \pm 2.5$ \\
 & & Left & $81.2 \pm 2.4$  \\
 & \multirow{2}{*}{Caucasian} &Left & $ 79.6$\\
  &  & Right & $  79.8$ \\
\hline\hline
\end{tabular}
\end{table*}

\begin{table*} %Ethnicity Prediction - males/females contrasting
\caption{Race prediction results for intra-gender and inter-gender training and testing (BioCOP2009 using 8-bit BSIF with a 9x9 filter).}
\label{raceMaleFemaleContrast}
\centering
\begin{tabular}{c c c c}
\hline\hline
\multicolumn{4}{c}{\textbf{Race}}\\
\hline\hline
\multirow{2}{*}{ \textbf{Train On}} &\multirow{2}{*}{ \textbf{Test On}} &\multirow{2}{*}{\textbf{Eye}}& \textbf{Prediction} \\
 &&&\textbf{Accuracy}\\
 \hline

 \multirow{4}{*}{Male}&  \multirow{2}{*}{Male} & Left & $ 92.9 \pm 2.1 $ \\
 &   &Right & $ 92.2 \pm 1.0 $ \\
 & \multirow{2}{*}{Female} &Left & $78.6$ \\
 & & Right & $78.7$\\
 \hline
 \multirow{4}{*}{Female} & \multirow{2}{*}{Female} &Right & $87.0 \pm 1.8 $ \\
 & & Left & $88.9 \pm 2.0 $  \\
 & \multirow{2}{*}{Male} &Left & $88.3$\\
  &  & Right & $88.3$ \\

\hline\hline
\end{tabular}
\end{table*}

\subsection{Impact of Image Blur on Race and Gender Prediction}
During the image acquisition process, ocular images may be captured out-of-focus. In order to determine the impact of out-of-focus images on both gender and race prediction, an additional experiment was performed. Out-of-focus images were simulated by `blurring' the image. The blurring effect was generated by applying a Gaussian filter to each image in the test partition with different sigma values ($\sigma$ = 2, 4, 6, 8, 10). Only the images in the test partition were blurred, while the images in the training partition were not blurred. The same subject-disjoint experimental protocol used in the previous sections was followed (see Section \ref{genderSubsection} and \ref{raceSubsection}).  The results are displayed in Table \ref{blurredImagePredictionAccuracy}.

Results from the experiment indicate that race prediction accuracy degrades at a steeper rate than gender prediction accuracy as the blurriness (i.e., sigma level) increases. We can conclude from this that race cues are at a much finer level than gender cues.

\begin{figure*}[t!]
\centering
\begin{subfigure}[t]{0.15\textwidth}
\centering
\includegraphics[height=0.8in]{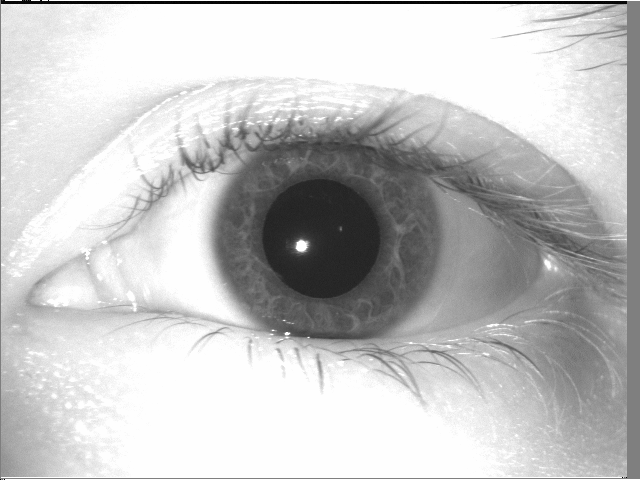}
\caption{Unmodified}
\end{subfigure}
\begin{subfigure}[t]{0.15\textwidth}
\centering
\includegraphics[height=0.8in]{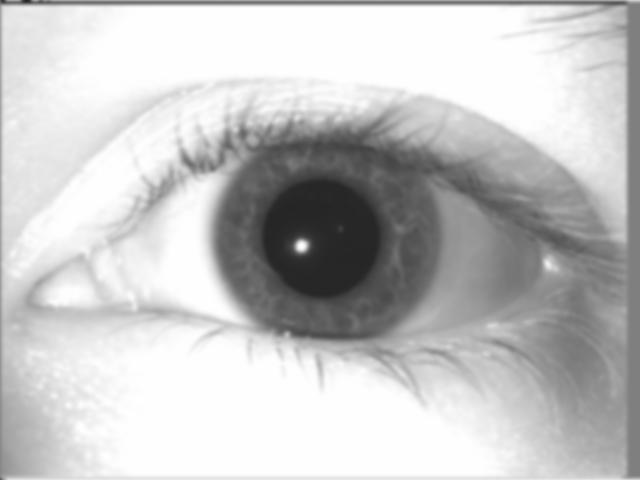}
\caption{$\sigma=2$}
\end{subfigure}
\begin{subfigure}[t]{0.15\textwidth}
\centering
\includegraphics[height=0.8in]{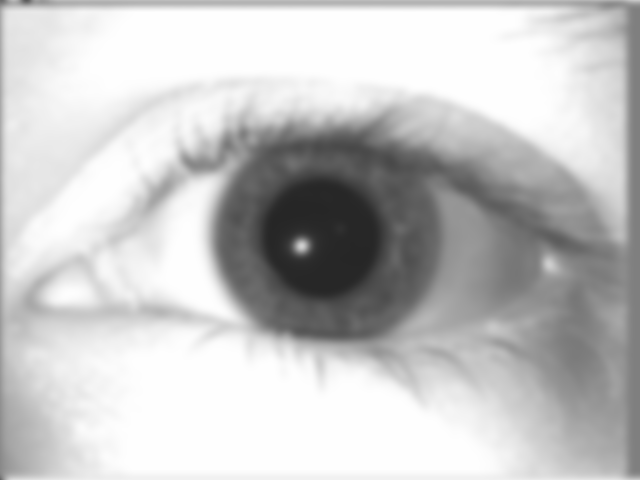}
\caption{$\sigma=4$}
\end{subfigure}
\begin{subfigure}[t]{0.15\textwidth}
\centering
\includegraphics[height=0.8in]{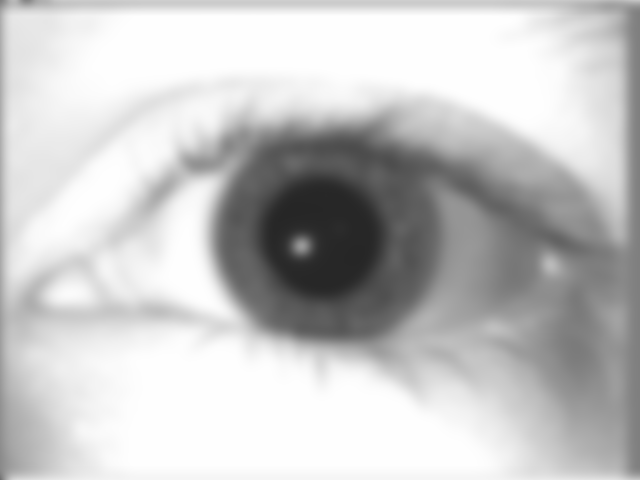}
\caption{$\sigma=6$}
\end{subfigure}
\begin{subfigure}[t]{0.15\textwidth}
\centering
\includegraphics[height=0.8in]{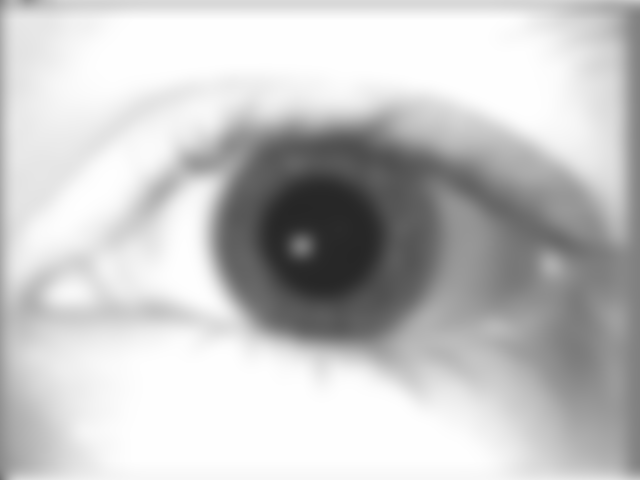}
\caption{$\sigma=8$}
\end{subfigure}
\begin{subfigure}[t]{0.15\textwidth}
\centering
\includegraphics[height=0.8in]{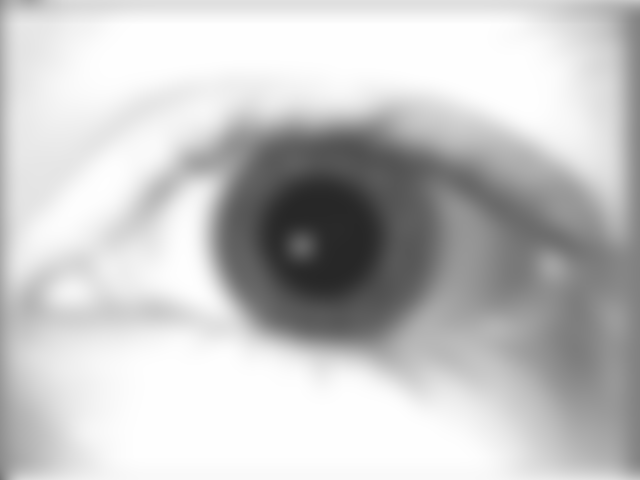}
\caption{$\sigma=10$}
\end{subfigure}
\caption{A sample ocular image that has been convolved with a Gaussian filter at different sigma values. The image in (a) is from \cite{doyle2013variation}.}
\label{sampleBlurredImages}
\end{figure*}

\begin{table*}
\caption{Gender and race prediction accuracy on blurred ocular images (BioCOP2009 using 8-bit BSIF with a 9x9 filter). Training is done on the original images in the train partition, while testing is done on the blurred images on the test partition.}
\label{blurredImagePredictionAccuracy}
\centering
\begin{tabular}{c |c c c c c c}
\hline\hline
&\multicolumn{6}{c}{\textbf{Attribute prediction accuracy from blurred images ($\%$) }}\\
\hline
%&\multicolumn{4}{c}{\textbf{Sigma}}\\
%&\multicolumn{4}{c}{\textbf{}}\\
% $\sigma=0$
Attribute&Unmodified&$\sigma=2$&$\sigma=4$&$\sigma=6$&$\sigma=8$&$\sigma=10$\\
%\multirow{2}{*}{\textbf{Attribute}}&Unmodified&$\sigma=2$&$\sigma=4$&$\sigma=6$&$\sigma=8$&$\sigma=10$\\
%&&&&&&\\
\hline
Gender&$85.9\pm0.7$&$83.1\pm0.7$ &$78.6\pm1.1$ &$75.3\pm1.7$ &$73.1\pm2.6$ &$70.8\pm3.2$ \\
Race&$89.8\pm1.5$&$83.1\pm1.5$&$64.8\pm2.2$ &$60.3\pm2.5$&$58.4\pm2.1$ &$57.2\pm2.0$\\
\hline
\end{tabular}
\end{table*}

\subsection{Impact of Eye Color on Race and Gender Prediction}
Certain eye colors are predominant to specific races~\cite{sturm2009genetics}; therefore, it is important to investigate the impact of eye color on both race and gender prediction. The breakdown of eye color\footnote{As in most iris data collection activities, eye color was self-declared by the subject and visually confirmed by the data collector.} by ethnicity and gender for the BioCOP2009 dataset is listed in Table \ref{table:biocop09DatasetEthnicityGenderColorBreakdown}. The gender and race prediction accuracies categorized by eye color are shown in Tables \ref{eyeColorImpactGenderPrediction} and \ref{eyeColorImpactRacePrediction}. Understanding the impact of eye color on race and gender prediction may help in designing a fusion model for attribute prediction.

The results shown in Table \ref{eyeColorImpactGenderPrediction} suggest that eye color does not have a significant impact on gender prediction. Males slightly outperform females regardless of eye color as seen in Table \ref{genderOcularConfusionMatrix}. However, for race prediction, Table \ref{eyeColorImpactRacePrediction} suggests that eye color does impact accuracy. Caucasian eyes exhibit a higher race prediction accuracy when the eye color is not brown, while Non-Caucasians eyes exhibit a higher race prediction accuracy when the eye color is brown. This observation may be related to the high number of Non-Caucasians with brown eyes in the dataset and limited number of Non-Caucasians with blue, green or hazel eye colors. Non-Caucasian subjects with light eye color is not just a characteristic of the BioCOP2009 dataset - Sturm and Larsson \cite{sturm2009genetics} have reported that light eye colors (blue, green, hazel) are found more frequently in Caucasians.

\begin{table*}
\caption{Eye color statistics by ethnicity and gender for the BioCOP 2009 dataset.}
\label{table:biocop09DatasetEthnicityGenderColorBreakdown}
\centering
\begin{tabular}{c|| c c |c  c || c c |c c}
\hline
\multirow{2}{*}{\textbf{Eye Color}}&\multicolumn{2}{c}{\textbf{Caucasian}}&\multicolumn{2}{c}{\textbf{Non-Caucasian}}&\multicolumn{2}{c}{\textbf{Male}}&\multicolumn{2}{c}{\textbf{Female}}\\
&Subjects&Images&Subjects&Images&Subjects&Images&Subjects&Images\\
\hline
Brown&267&10,330    &228&8470      &235&8,983     &260&9,817\\
Blue&294 &11,157    &2 &87              &119&4,638     &177 &6,606\\
Green&137 &5,251    &6 &226   &46 &1,806          &97 &3,671\\
Hazel&130 &5,055    &6 &226         &50 &1,955          &86 &3,326\\
Gray&8 &294            &0& 0           &3 &119         &5&175\\
Other&0 &0          &18 &734            &14 &543       &4 &191\\
\hline
\end{tabular}
\end{table*}

\begin{table*}
\caption{Impact of eye color on gender prediction (BioCOP2009 using 8-bit BSIF with a 9x9 filter).}
\label{eyeColorImpactGenderPrediction}
\centering
\begin{tabular}{l| c| c| c |c}
\hline\hline
\multicolumn{5}{c}{\textbf{Gender Prediction Accuracy (\%)}}\\
\hline\hline
%&\multicolumn{2}{c}{\textbf{Left}}&\multicolumn{2}{c}{\textbf{Right}}\\
&\multicolumn{2}{c}{ \textbf{Male}} &\multicolumn{2}{c}{ \textbf{Female}} \\
\hline
\textbf{Eye Color}&\textbf{Left}&\textbf{Right}&\textbf{Left}&\textbf{Right}\\
\hline
Brown       & $86.8\pm 3.4$ &  $87.0\pm2.33$       & $79.0\pm 3.4$ & $81.0\pm2.2$ \\
Blue            & $92.2\pm 1.1$ &  $84.6\pm1.60$       &$84.7\pm3.9$ &  $85.2\pm1.5$ \\
Green         & $93.0\pm 1.1$&   $91.2\pm2.15$      &$84.7\pm3.5$ & $88.9\pm2.4$\\
Hazel          & $87.1\pm2.5$& $85.0\pm3.9$            &$91.0\pm2.6$ &  $85.7\pm2.8$\\
\hline
\end{tabular}
\end{table*}

\begin{table*}
\caption{Impact of eye color on race prediction (BioCOP2009 using 8-bit BSIF with a 9x9 filter).}
\label{eyeColorImpactRacePrediction}
\centering
\begin{tabular}{l| c| c| c |c}
\hline\hline
\multicolumn{5}{c}{\textbf{Race Prediction Accuracy (\%)}}\\
\hline\hline
%&\multicolumn{2}{c}{\textbf{Left}}&\multicolumn{2}{c}{\textbf{Right}}\\
&\multicolumn{2}{c}{ \textbf{Caucasian}} &\multicolumn{2}{c}{ \textbf{NonCaucasian}} \\
\hline
\textbf{Eye Color}&\textbf{Left}&\textbf{Right}&\textbf{Left}&\textbf{Right}\\
\hline
Brown& $79.1\pm3.2$ & $83.6\pm2.9$     & $90.4\pm1.2$ & $90.5\pm1.5$ \\
Blue& $98.5\pm1.1$ & $95.5\pm 0.6$          & $0.0\pm0.0$ &$1.8\pm2.2$\\
Green&$99.6\pm0.3$ & $90.2\pm3.1$          & $20.4\pm11.7$ & $6.0\pm10.8$ \\
Hazel&$90.1\pm2.7$& $90.5\pm3.6$          & $64.1\pm45.6$ & $46.4\pm34.2$ \\
%Other& $NaN$ & $NaN$                     & $76.3\pm1.1$&$85.8\pm1.5$\\
    %$ 92.9 \pm 2.1 $
\hline
\end{tabular}
\end{table*}

\subsection{Texture Descriptor Comparison}
In order to select a suitable texture descriptor for experiments in this work, three were first considered: BSIF, LBP and LPQ. Each of the three texture descriptors are described in Section~\ref{featureExtractionSection}. Prediction accuracies were generated using the proposed methods from Section~\ref{raceSubsection} and \ref{genderSubsection} for race and gender, respectively. The results of these experiments are shown in Tables~\ref{raceTextureDescriptorComparison} and \ref{genderTextureDescriptorComparison}. BSIF was selected as the primary texture descriptor based on it's overall performance.

\section{Discussion}
\label{discussionSection}

In this paper, a number of experiments were performed to provide insight into the problem of predicting race and gender from NIR ocular images. Our broad findings are summarized below:

\begin{itemize}

    \item \textbf{Texture Descriptors:} Gender and race prediction can be accomplished using simple texture descriptors. Both gender and race are predicted using the same feature vector (see Tables \ref{genderIrisOnlyIrisExcludedResults} and \ref{raceTextureDescriptorComparison}).

    \item \textbf{Generalizability:} The proposed algorithm is generalizable across multiple datasets and is, therefore, learning more than just artifacts from a single dataset. The generalizability applies for both gender and race prediction (see Tables~\ref{raceCrossDatasetTesting} and  \ref{genderCrossDatasetTesting}).

    \item \textbf{Gender prediction:} The iris-excluded region provides greater prediction accuracy for gender than the iris-only region (see Table~\ref{genderIrisOnlyIrisExcludedResults}).

    \item \textbf{Race prediction:} The iris-only region provides greater prediction accuracy for race than the iris-excluded region when utilizing BSIF as the texture descriptor (see Table~\ref{irisExcludedIrisOnlyRace}).

   \item \textbf{Left and Right:} There is no significant difference in performance between the left and right eye images for gender and race prediction (see Table \ref{genderCrossDatasetTesting}).

    \item \textbf{Cross-gender training: }For race prediction, training only on \emph{male} images and testing on only \emph{female} images results in a $\sim$14\% decrease in prediction accuracy than when training and testing on only male images. Training only on \emph{female} images (also for race prediction) and testing on only \emph{male} images shows no significant difference in prediction accuracy (see Table \ref{raceMaleFemaleContrast}).

    \item \textbf{Region-based descriptor performance:}  For race prediction, LBP outperformed BSIF for the iris-excluded region, while BSIF outperformed LBP for the iris-only and extended ocular region (see Table~\ref{raceTextureDescriptorComparison}).

     \item \textbf{Impact of eye color on race and gender prediction:} For race prediction, Non-Caucasians with brown eyes displayed a higher prediction accuracy than Caucasians with brown eyes (see  Table~\ref{eyeColorImpactRacePrediction}). For gender prediction there was no observable impact based on eye color (see Table~\ref{eyeColorImpactGenderPrediction}).

     \item \textbf{Impact of image blur on race and gender prediction:} The prediction accuracy for race degrades at a much faster rate than gender as the $\sigma$ value of the Gaussian filter for blurring is increased (see Table~\ref{blurredImagePredictionAccuracy}).

\end{itemize}

\section{Future Work}
\label{futureWorkSection}

In future work, the number of predicted attributes could be expanded beyond just race and gender. Age, eye color~\cite{bobeldyk2018predicting} and texture smoothness  are just a few of the other attributes that could be explored. Investigating the correlations between these attributes would be essential in better understanding the relationship between them and their role in iris recognition performance.

Previous work has shown \cite{jain2004can} that the fusion of soft biometric information with a traditional biometric recognition system can increase the overall recognition accuracy of the system. Developing such a fusion framework for iris recognition based on the extracted attributes may result in improved performance in non-ideal scenarios.

BSIF outperforms other texture descriptors in most of the experiments presented in this paper. Future research could explore why BSIF outperforms the other texture descriptors in many of these experiments. Fusing the outputs of multiple texture descriptors and/or a convolutional neural network may result in further improvement in performance. Specifically, utilizing different descriptors on different ocular regions may result in improved attribute prediction. Finally, it may be possible to fuse the BSIF feature vector with the iris code to increase recognition performance and attribute prediction, simultaneously.

\flushend
\newpage

\bibliography{../../../../bibliography/irisGender}

\newpage
\begin{IEEEbiography}
[{\includegraphics[width=1.1in,height=1.75in,clip,keepaspectratio]{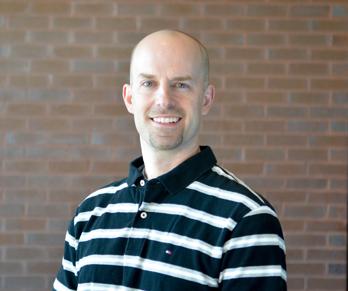}}]
Denton Bobeldyk received the B.S. degree in Computer Science from Calvin College, Grand Rapids, United States, in 1995, and the M.S. from Grand Valley State University, Allendale in 2004. In 2008 he accepted a job at Davenport University, where he is currently an Associate Professor. He worked on a team to develop their Computer Science undergraduate and graduate programs. In 2013, he joined the PhD program at Michigan State University and is currently studying under Dr. Arun Ross in the iProbe research lab. He is the recipient of the Davenport University 2014 Excellence in Teaching Award and was a finalist for the same award in 2018. \end{IEEEbiography}

\begin{IEEEbiography}
[{\includegraphics[width=1in,height=1.5in,clip,keepaspectratio]{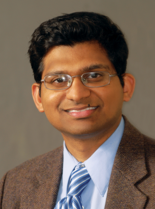}}]
Arun Ross received the B.E. (Hons.) degree in computer science from the Birla Institute of Technology and Science, Pilani, Inida, in 1996, and the M.S. and Ph.D. degrees in computer science and engineering from Michigan State University, East Lansing, in 1999 and 2003, respectively. He has authored the books \emph{Introduction to Biometrics: A Textbook} and \emph{Handbook of Multibiometrics}, and coedited the book \emph{Handbook of Biometrics}. His research interests include pattern recognition, classifier fusion, computer vision, and biometrics. He is a recipient of the NSF's CAREER Award and was designated a Kavli Frontier Fellow by the National Academy of Sciences in 2006. He is currently a Professor with the Department of Computer Science and Engineering, Michigan State University.
He is a senior area editor of the IEEE Transactions on Image Processing; associate editor of the Image and Vision Computing Journal and ACM Computing Surveys; and Area Editor of the Computer Vision and Image Understanding Journal.
\end{IEEEbiography}

%\section*{References}

%\bibliography{mybibfile}

\end{document}